\documentclass[journal]{IEEEtran}

\ifCLASSINFOpdf
\else
\fi
\usepackage{times}
\usepackage{xcolor}
\usepackage{soul}
\usepackage[utf8]{inputenc}
\usepackage[small]{caption}
\usepackage{multirow}
\usepackage{graphicx}  
\usepackage{helvet}
\usepackage{courier}
\usepackage{url}
\usepackage{color}
\usepackage{amsmath}
\usepackage{amssymb}
\usepackage[ruled,vlined]{algorithm2e}

\newtheorem{myDef}{Definition}

\usepackage[loose]{subfigure}
\usepackage{epstopdf}

\usepackage{float}
\usepackage[T1]{fontenc}

\usepackage{enumerate}

\def\eg{{\em e.g.}}
\def\ie{{\em i.e.}}
\def\etal{{\em et al.}}

\begin{document}
%
\title{Multi-View Deep Subspace Clustering Networks}
%

\author{Pengfei Zhu, Xinjie Yao, Yu Wang,~\IEEEmembership{Member,~IEEE}, Binyuan Hui, Dawei Du, Qinghua Hu,~\IEEEmembership{Senior~Member,~IEEE}

\thanks{Pengfei Zhu, Xinjie Yao, Yu Wang, Binyuan Hui, and Qinghua Hu are with the College of Intelligence and Computing, Tianjin University, Tianjin 300350, China (e-mail: zhupengfei$@$tju.edu.cn; yaoxinjie$@$tju.edu.cn; wang.yu$@$tju.edu.cn; huibinyuan$@$tju.edu.cn; huqinghua$@$tju.edu.cn). Pengfei Zhu, Xinjie Yao, Yu Wang, and Qinghua Hu are also with the Engineering Research Center of City Intelligence and Digital Governance, Ministry of Education of the People's Republic of China, and Haihe Lab of ITAI.

Dawei Du is with the Computer Science Department, University at Albany, State University of New York, Albany, NY, USA (e-mail: ddu@albany.edu).

Yu Wang is the corresponding author.

This work was supported in part by the National Key R\&D Program of China under Grant 2022ZD0116500, in part by the National Natural Science Foundation of China under Grants 62106174, 62222608, 62266035, and 61925602, and in part by Tianjin Natural Science Funds for Distinguished Young Scholar under Grant 23JCJQJC00270.
}}

\markboth{IEEE TRANSACTIONS ON CYBERNETICS}%
{Shell \MakeLowercase{\textit{et al.}}: Bare Demo of IEEEtran.cls for IEEE Journals}

\maketitle

\begin{abstract}
Multi-view subspace clustering aims to discover the inherent structure of data by fusing multiple views of complementary information.
Most existing methods first extract multiple types of handcrafted features and then learn a joint affinity matrix for clustering.
The disadvantage of this approach lies in two aspects: 1) multi-view relations are not embedded into feature learning, and 2) the end-to-end learning manner of deep learning is not suitable for multi-view clustering.
Even when deep features have been extracted, it is a nontrivial problem to choose a proper backbone for clustering on different datasets.
To address these issues, we propose the Multi-view Deep Subspace Clustering Networks (MvDSCN), which learns a multi-view self-representation matrix in an end-to-end manner.
The MvDSCN consists of two sub-networks, \ie, a diversity network (Dnet) and a universality network (Unet).
A latent space is built using deep convolutional autoencoders, and a self-representation matrix is learned in the latent space using a fully connected layer.
Dnet learns view-specific self-representation matrices, whereas Unet learns a common self-representation matrix for all views.
To exploit the complementarity of multi-view representations, the Hilbert--Schmidt independence criterion (HSIC) is introduced as a diversity regularizer that captures the nonlinear, high-order inter-view relations.
Because different views share the same label space, the self-representation matrices of each view are aligned to the common one by universality regularization.
The MvDSCN also unifies multiple backbones to boost clustering performance and avoid the need for model selection.
Experiments demonstrate the superiority of the MvDSCN.
\end{abstract}


\section{Introduction}
Subspace clustering aims to segment a set of unlabeled samples drawn from a union of multiple subspaces corresponding to different clusters into several groups.
Recently, self-representation based models have achieved state-of-the-art (SOTA) performance in subspace clustering~\cite{Guangcan2013Robust,ji2017deep,rw_subspace_tcyb_2}.
Given data points $\{x_i\}_{i=1,..., N}$ drawn from multiple linear subspaces $\{S_i\}_{i=1,..., K}$, one can express a point in a subspace as a linear combination of other points in the same subspace. In the literature \cite{ji2017deep}, this property is called self-expressiveness or self-representation. If we stack all the points $x_i$ into columns of a data matrix ${\bf{X}}$, the self-expressiveness property can be simply represented as one single equation, \ie, ${\bf{X}} ={\bf{XC}}$, where ${\bf{C}}$ is the self-representation coefficient matrix. Self-representation assumes that a sample can be represented by a linear combination of a set of samples
\begin{equation}\label{sr}
  \begin{array}{l}
\mathop {\min }\limits_{\bf{Z}} L({\bf{X}},{\bf{Z}}) + R({\bf{Z}}),\; \; s.t.\quad{\bf{X}} = {\bf{XZ}},
\end{array}
\end{equation}
where $\bf{X} \in \mathbb{R}^{d \times n}$ and $\bf{Z} \in \mathbb{R}^{n \times n}$ denote the training data and self-representation matrix, respectively.
Furthermore, $L({\bf{X}},{\bf{Z}})$ represents the reconstruction loss and $R({\bf{Z}})$ is the regularization item.
The key differences in self-representation based subspace clustering models lie in the option of the loss function and regularizer.
For $R({\bf{Z}})$, the $l_0$-norm, $l_1$-norm, square of the Frobenius norm, elastic net, trace Lasso, and k-block diagonal regularizers have been used under certain subspace assumptions \cite{Lu2018Subspace}.
Because handcrafted features cannot capture extreme variations well, deep subspace clustering models have been developed to jointly learn hierarchical representation~\cite{rw_wang2021hierarchical_tcyb_6} and cluster structure \cite{rw_subspace_tcyb_2,Jiang2018WhenTL}.

\begin{figure}
	\centering
	\includegraphics[width=1\linewidth]{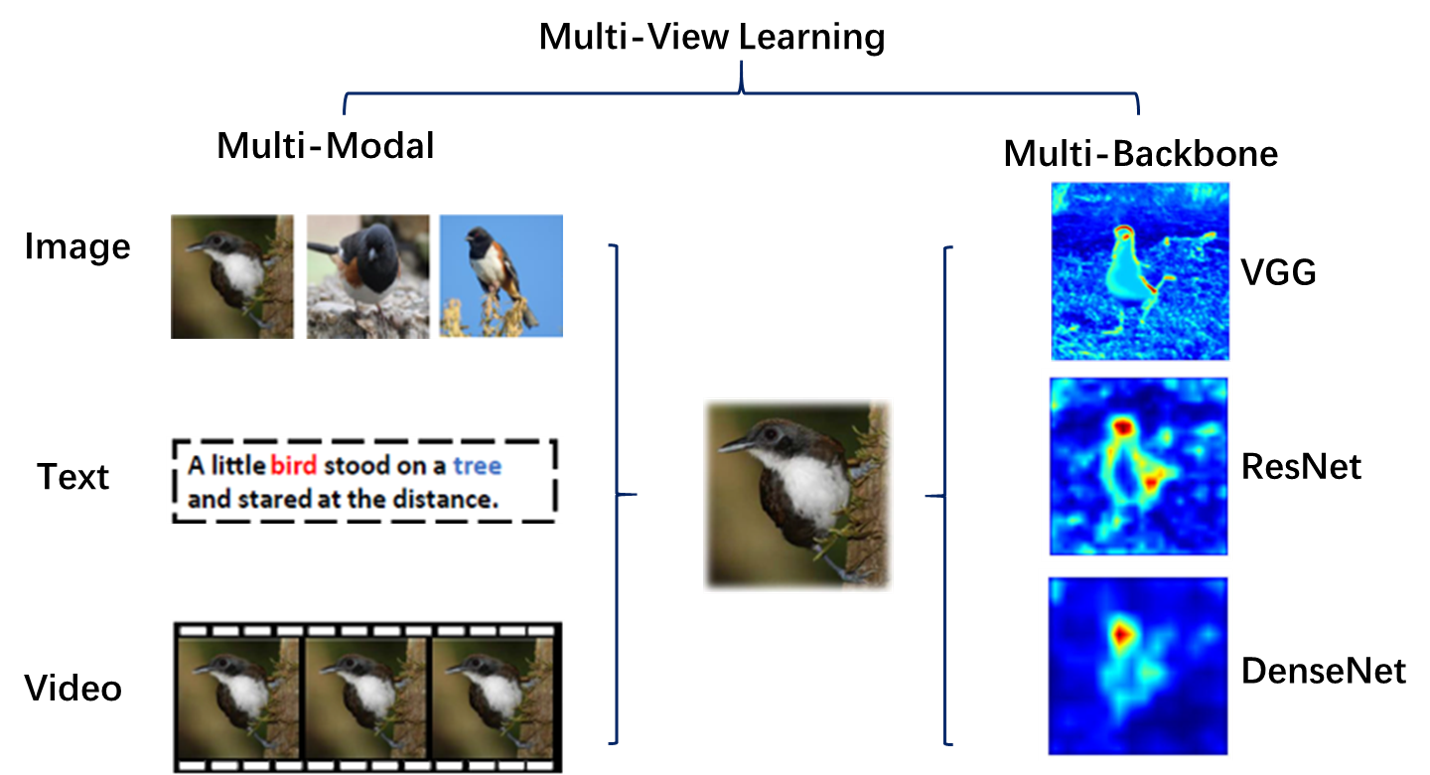}
	\caption{
{Examples of multi-view learning.
A sample can be represented by different modalities, such as images, video, and text.
Different kinds of backbones, \eg, VGG, ResNet, and DenseNet, can be used to obtain multi-view feature representation}. }
	\label{mv}
\end{figure}

The rapid growth of digital sensors and widespread application of social networks has led to an explosion of multi-modal data in areas such as multimedia analysis \cite{luo2017simple}, medical image analysis \cite{kim2015dasc}, and autonomous driving \cite{chen2017multi}.
As shown in Fig. \ref{mv}, different types of data can be collected, including text, image, audio, and video data, to represent a sample.
{
Even for single modal data, \eg, images or video sequences, diverse features can be extracted to capture the scale, occlusion, illumination, and rotation variations for robust recognition \cite{Zhao2017MultiViewCV}.
Multi-modal and multi-feature information can be fused to boost the performance of subspace clustering}.
Besides, because there are different types of backbones for deep neural networks, \eg, VGG, ResNet, and DenseNet, it is still a nontrivial problem to select a proper backbone for clustering on different datasets.
If we can combine multiple backbones into a unified model, the clustering performance can be boosted by the fusion of multiple backbones.

Multi-view subspace clustering (MvSC) aims to use data collected from different modalities or represented by different types of features to discover the underlying clustering structure \cite{li2021self,zhou2019dual,tang2021one}.
Most MvSC methods design a multi-view regularizer to characterize the inter-view relationships between several types of handcrafted features for multi-view clustering~\cite{Zhao2017MultiViewCV,xing2019correntropy,chen2020multiview,Zhang2017LatentMS}.
However, their performances are still not satisfactory for two reasons.
First, existing methods adopt a two-stage strategy, \ie, extracting features and then learning the affinity matrix. The feature extraction process is not relevant to the subspace clustering task.
Multi-view relationships can only be considered during the affinity matrix learning process, which ignores the role of inter-view relations in feature learning.
Second, they consider little about hierarchical representation learning in an end-to-end manner.

In this paper, we propose the Multi-view Deep Subspace Clustering Networks (MvDSCN), which learns a multi-view self-representation matrix in an end-to-end manner. In summary, our contributions to this work are summarized as follows:
\begin{enumerate}[1)]
\item We propose a novel multi-view clustering method termed MvDSCN via deep subspace. The MvDSCN unifies multiple backbones and combines convolutional autoencoders with self-representation to learn multi-view self-representation in an end-to-end manner.
\item The explicitly decoupled representation learning MvDSCN consists of two sub-networks, \ie, a diversity network (Dnet) and a universality network (Unet), that embed multi-view relations into feature learning to perform better clustering. 
\item With diversity and universality regularization, multi-view relations are well embedded in feature learning and self-representation stages. Experiments on multi-view clustering show the effectiveness of the proposed model with respect to SOTA subspace clustering methods.
\end{enumerate}

The rest of this paper is organized as follows.
Section \ref{s2} introduces the related work on multi-view learning, self-representation, and autoencoders.
Section \ref{s3} describes the proposed multi-view clustering model.
Section \ref{s4} reports the results of experiments on multi-view clustering tasks.
Section \ref{s5} concludes the paper with a discussion of future work.

\section{Related Work}
\label{s2}
\subsection{Multi-view Clustering}
Subspace clustering aims to uncover the inherent cluster structure from data composed of multiple subspaces \cite{rw_subspace_tcyb_2,rw_subspace_tcyb_3}.
In the past few years, most existing subspace clustering methods have focused on learning the affinity matrix and conducting spectral clustering.
Self-representation based subspace clustering methods are essentially based on the hypothesis that a sample can be reconstructed by a linear combination of other samples.
The low-rank representation (LRR) method explores the multi-block diagonal property of the self-representation matrix to discover multiple subspace structures \cite{Guangcan2013Robust,Lu2018Subspace,rw_low_rank_tcyb_1,rw_low_rank_conference_1}.
A deep subspace clustering network embeds self-representation into a deep convolutional autoencoder through a fully connected layer \cite{ji2017deep}.

Multi-view learning boosts clustering performance by exploring complementary information by modeling inter-view relations or learning a latent representation \cite{yu2020clustering,rw_mv_survey_1}.
{To model the problem of multi-view relations, MVGL shares the latent factors from all views. In addition, the graph construction and label propagation procedures can be jointly optimized for semi-supervised multi-view learning \cite{Li2017MultiviewGL}}.
For multi-view clustering, most existing methods can be considered to be an extension of single-view models, including spectral clustering \cite{chen2021multiview,rw_spectral_tcyb_5}, matrix factorization \cite{Zhao2017MultiViewCV}, and k-means \cite{kang2021structured}.
{The multi-view ensemble clustering (MVEC) algorithm learns a common low-rank representation shared by multi-view partitions \cite{Tao2017FromEC}.
	Furthermore, the M2VEC model incorporates a marginalized denoising autoencoder (mDA) into the ensemble model to boost the	robustness and extends MVEC to a multi-layer architecture \cite{Tao2019MarginalizedME}.}
Multi-view relations can be generally categorized into universality and diversity relations \cite{Zhang2017LatentMS,chen2021multiview,kang2021structured}. Contrastive learning has offered novel insights into achieving a unified representation of diverse views while simultaneously addressing view consistency and instance completeness \cite{DCP,SURE}. The tensor approach for uncoupled multi-view clustering (T-UMC) addresses challenges posed by uncoupled data through view-specific silhouette coefficients and tensor-based techniques~\cite{rw_tensor_tcyb_4}. Global and cross-view feature aggregation for multi-view clustering (GCFAggMVC) innovatively combines cross-view and cross-sample feature aggregation with a structure-guided contrastive learning module~\cite{rw_aggregation_conference_2}.

Universality emphasizes that all views should be similar, whereas diversity focuses on inter-view complementarity, which induces a diverse, view-specific representation.
Some works build multi-view connections by a common latent representation for clustering and model multi-view relations using neural networks \cite{ZhangGeneralized}.
Deep learning has performed excellently on many tasks because of its end-to-end learning approach.
However, existing multi-view subspace clustering methods treat multi-view feature extraction and affinity learning as separate stages.
In addition, because of the view-specific characteristic, it is difficult to force the self-representation matrices of all views to be the same.

\subsection{Self-representation}
Self-representation reflects the intra-relations among samples and has been widely used in image processing, clustering, feature selection, and deep learning.
In image processing, especially image denoising, the non-local mean has been widely used by reconstructing a pixel or image patch using related pixels or patches in the image \cite{buades2005non}, inspiring many successful image processing models in low-level vision.
As an alternative to pixel-level self-representation, a sample can be reconstructed well by a linear combination of bases.
Self-representation has been successfully used for clustering because it can accurately capture the sample relations by embedding sparse, dense, or low-rank priors \cite{Guangcan2013Robust,ji2017deep,Lu2018Subspace}.
To counter the curse of dimensionality, feature selection aims to select a subset of features by evaluating their importance.
Feature-level self-representation assumes that one feature can be reconstructed by all features, and the self-representation coefficients can be used for feature evaluation \cite{zhu2017subspace}.
Inspired by the success of non-local mean in image denoising, a non-local neural network is proposed to use the relations across elements of feature maps, channels, or frames to improve the representation ability of the networks \cite{wang2018non}.

\subsection{Autoencoders}
Autoencoders extract features of data by mapping the data to a low-dimensional space.
With the rapid development of deep learning, deep (or stacked) autoencoders have become popular for unsupervised learning.
Deep autoencoders have been widely used in dimensionality reduction \cite{Hinton2006ReducingTD} and image denoising \cite{Vincent2010StackedDA}.
Recently, deep autoencoders have been used to initialize deep embedding networks for unsupervised clustering \cite{Xie2016UnsupervisedDE}.
The work in \cite{Peng2016DeepSC} uses fully connected deep autoencoders by incorporating a sparsity prior into the hidden representation learning to preserve the sparse reconstruction relation.
By contrast, Ji \etal \cite{ji2017deep} directly learn the affinities between all data points through a deep autoencoder network using a fully connected self-representation layer.

Convolutional layers have fewer parameters but a stronger learning ability than fully connected layers, and convolutional autoencoders (CAEs) that can be trained in an end-to-end manner have been designed to learn features from unlabeled data.
Convolutional neural networks can be initialized by a CAE stack \cite{Masci2011StackedCA}, which is an unsupervised method for hierarchical feature extraction.
CAEs have also been successfully used for generative adversarial networks (GANs).
Nguyen \etal \cite{Nguyen2017PlugP} combines a convolutional autoencoder loss, a GAN loss, and a classification loss defined using a pre-trained classifier.

\begin{figure*}[!htbp]
	\centering
	\includegraphics[width=0.95\textwidth]{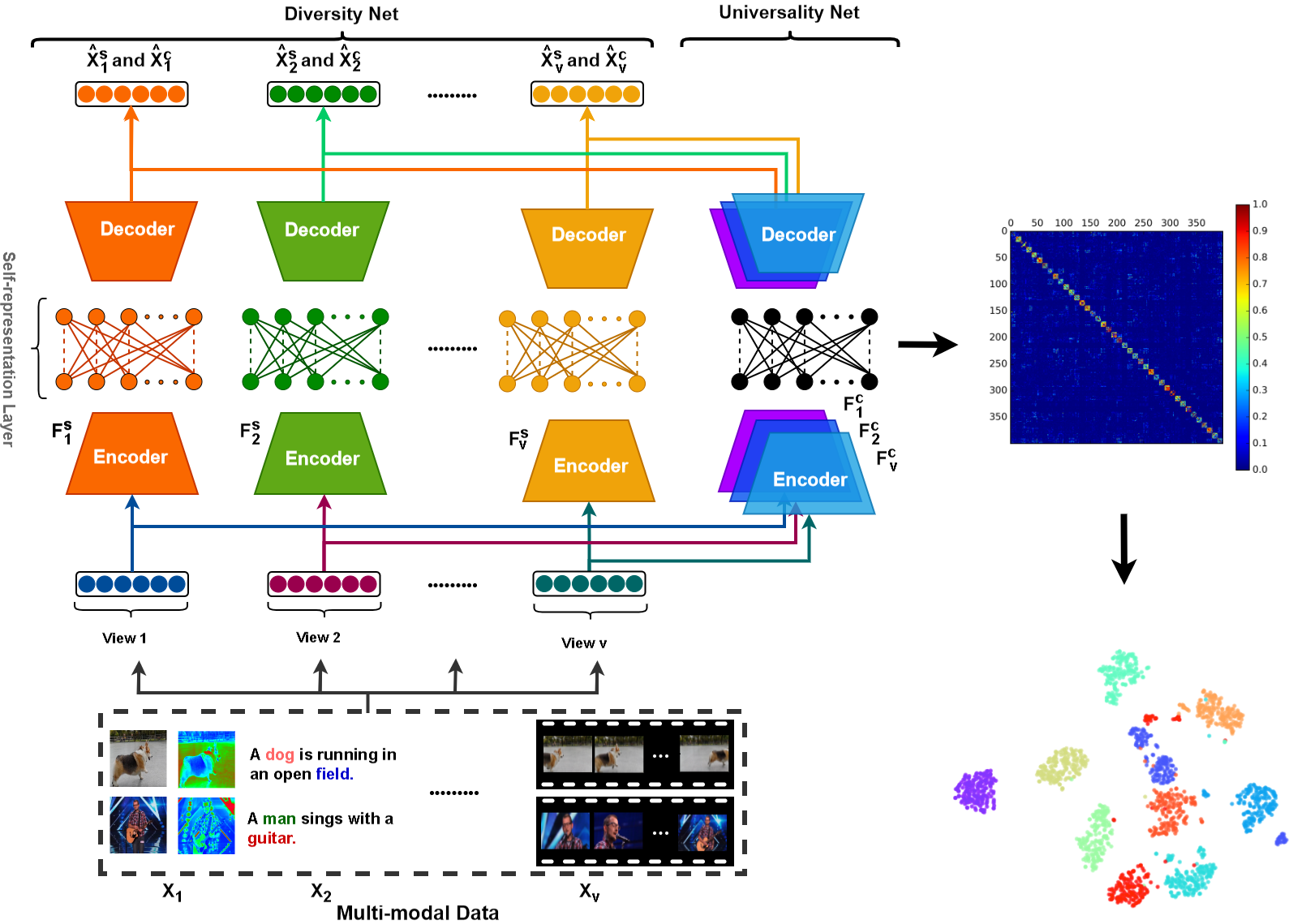}
	\caption{{MvDSCN, which consists of two parts, \ie, the Dnet, which learns view-specific representation using different autoencoders and an independent self-representation layer, and the Unet, which learns view-consistent representation using different autoencoders and a common self-representation layer.}}
	\label{architecture}
\end{figure*}

\section{Multi-view Deep Subspace Clustering}
\label{s3}
In this section, we provide a detailed description of the MvDSCN, with the framework illustrated in Fig. \ref{architecture}. The MvDSCN consists of two sub-networks: a Dnet and a Unet. A latent space is constructed using deep convolutional autoencoders, and a self-representation matrix is learned within this latent space through a fully connected layer. Dnet focuses on learning view-specific self-representation matrices, while Unet is responsible for acquiring a universal self-representation matrix applicable to all views. To leverage the complementary nature of multi-view representations, HSIC is introduced as a diversity regularizer. HSIC captures the nonlinear, high-order inter-view relationships. As different views share the same label space, the self-representation matrices of each view are harmonized with the common one through universality regularization.

\subsection{Network Architecture}
Let ${\bf{X}}_1,..., {\bf{X}}_i,...,{\bf{X}}_v$ denote the inputs of multiple views, where ${\bf{X}}_i \in \mathbb{R}^{n \times d_i}$, $v$, $n$ and $d_i$ are the number of views, samples, and dimension of features in the $i$-th view, respectively.
$ {\bf{X}}_i$ are handcrafted features or raw data, such as image or RGB-D data.
The proposed network consists of two parts, \ie, the Dnet, which learns view-specific representations, and the Unet, which shares a view-consistent self-representation matrix.

The Dnet embeds the input ${\bf{X}}_i$ into a hidden representation ${\bf{F}}^s_i$ using the view-specific encoder for the $i^{th}$ view.
Then, self-representation is conducted by a fully connected layer without bias and nonlinear activations, \ie, ${\bf{F}}^s_i$=${\bf{F}}^s_i {\bf{Z}}_i$.
The Unet uses a common self-representation matrix $\bf{Z}$ for all views, which is connected with a hidden representation ${\bf{F}}^c_1, {\bf{F}}^c_2,..., {\bf{F}}^c_v$ of all views.
After the self-representation layer, the samples are recovered by the view-specific decoders. We hope to maintain consistency while considering the complementarity of views. Using a common representation ${\bf{Z}}$ for different ${\bf{\hat X}}_i^c$ of different views will help MvDSCN to maintain consistency.

For both the Dnet and Unet, we advocate using convolutional autoencoders, which have fewer parameters and stronger learning ability than fully connected layers.
We use three-layer encoders with $\{64, 32, 16\}$ channels, and three-layer decoders with $\{16, 32, 64\}$ channels.
We adopt a $3 \times 3$ kernel and rectified linear unit (ReLU) \cite{Nair2010RectifiedLU} for the nonlinear activations.
Notably, no pooling layers are used. The latent features are then restored to a space the same size as the input space via the transpose convolution layers.

\subsection{Loss Function}
The losses of the MvDSCN consist of two parts, \ie, the reconstruction loss and the self-representation loss. The reconstruction loss can be calculated as
\begin{equation}\label{loss_autoencoder}
\resizebox{.9\hsize}{!}{$
{L_{r}}({{\bf{X}}_i}, {\bf{\hat X}}_i^s, {\bf{\hat X}}_i^c) = 
\min \left( \begin{array}{l}
 \sum\limits_{i = 1}^v {\left\| {{{\bf{X}}_i} - {\bf{\hat X}}_i^s} \right\|_F^2 + \left\| {{{\bf{X}}_i} - {\bf{\hat X}}_i^c} \right\|_F^2}\end{array}\right)$},
\end{equation}where ${\bf{X}}_i$, ${\bf{\hat X}}_i^s$, and ${\bf{\hat X}}_i^c$ denote handcrafted features or raw data of the $i$-th view, data decoded by the view-specific decoder of the $i$-th view, and data decoded by the view-consistent decoder of the $i$-th view, respectively. The self-representation loss can be calculated as
\begin{equation}\label{loss_representation}
\resizebox{.9\hsize}{!}{$
{L_{s}}({\bf{F}}_i^s, {\bf{F}}_i^c, {{\bf{Z}}_i}, {\bf{Z}}) = 
\min \left( \begin{array}{l}
 \sum\limits_{i = 1}^v {\left\| {{\bf{F}}_i^s - {\bf{F}}_i^s{{\bf{Z}}_i}} \right\|_F^2 + \left\| {{\bf{F}}_i^c - {\bf{F}}_i^c{\bf{Z}}} \right\|_F^2}
 \end{array}\right)$},
\end{equation}where ${\bf{F}}_i^s$ and ${\bf{F}}_i^c$ are data encoded by the view-specific and view-consistent encoders of the $i$-th view, respectively; ${\bf{Z}}_i$ is the self-representation matrix of the $i$-th view; and ${\bf{Z}}$ is the common self-representation matrix. 

In order to represent the subspace more accurately, we use the $l_{p}$-norm regularizer, which can be defined as
\begin{equation}\label{regularizer_lp_norm}
R_{l_p}({{\bf{Z}}},{{\bf{Z}}_i}) = {{\left\| {\bf{Z}} \right\|_p^{}{\rm{ + }}\sum\limits_{i = 1}^v {\left\| {{{\bf{Z}}_i}} \right\|_p^{}} }}.
\end{equation}

Because all views share the same decision space, the view-specific self-representation matrices that reflect sample relations should be aligned with the common self-representation matrix used in Unet. In our framework, the structures of the encoders and decoders are the same, respectively. Therefore, different views are aligned. Further, we align the different views with a common self-representation matrix. As different views share the same latent space, the self-representation matrices of each view are aligned to the common one by a universality regularization, which is defined as follows:

\begin{equation}\label{Cengru}
{R_u}({{\bf{Z}}},{{\bf{Z}}_i}) = \sum\limits_{i = 1}^v {\left\| {{\bf{Z}} - {{\bf{Z}}_i}} \right\|_F^2}.
\end{equation}

To embed multi-view relations into feature learning and self-representation, two types of regularizers are used.
To exploit the complementary information from multiple views, \eg, RGB and depth information, the diversity regularization is defined based on the HSIC \cite{gretton2005measuring}.
The HSIC measures the nonlinear and high-order correlations and has been successfully used in multi-view subspace clustering.

Assuming that there are two variables ${\bf{A}} = \left[ {{{\bf{a}}_1},...,{{\bf{a}}_i},...,{{\bf{a}}_N}} \right]$ and ${\bf{B}} = \left[ {{{\bf{b}}_1},...,{{\bf{b}}_i},...,{{\bf{b}}_N}} \right]$,
we define a mapping $\phi ({\bf{a}})$ from ${\bf{a}} \in {\mathfrak{A}}$ to kernel space $\mathfrak{F}$, where the inner product of two vectors is defined as $k({{\bf{a}}_1},{{\bf{a}}_2}) = \left\langle {\phi ({{\bf{a}}_1}),\phi ({{\bf{a}}_2})} \right\rangle$.
Then, $\varphi ({\bf{b}})$ is defined to map ${\bf{b}} \in {\mathfrak{B}}$ to kernel space $\mathfrak{G}$.
Similarly, the inner product of two vectors in $\mathfrak{G}$ is defined as $g({{\bf{b}}_1},{{\bf{b}}_2}) = \left\langle {\phi ({{\bf{b}}_1}),\phi ({{\bf{b}}_2})} \right\rangle$.
The empirical version of HSIC is induced as follows.
\begin{myDef}
Consider a series of $N$ independent observations drawn from $p_\mathbf{ab}$, $\mathfrak{Z}:=\{(\mathbf{a}_1,\mathbf{b}_1),...,(\mathbf{a}_N,\mathbf{b}_N)\} \subseteq \mathfrak{A} \times \mathfrak{B}$. An estimator of HSIC, written as HSIC($\mathfrak{Z},\mathfrak{F},\mathfrak{G}$), is given by
\begin{equation}{
\text{HSIC}(\mathfrak{Z},\mathfrak{F},\mathfrak{G}) =  (N-1)^{-2}tr(\mathbf{G}_1\mathbf{H}\mathbf{G}_2\mathbf{H}),
}
\end{equation}
where $tr(\cdot)$ is the trace of a square matrix. In addition, $\mathbf{G}$ is the Gram matrix \cite{gatys2015neural}, expressed as follows:
\begin{equation}\label{Gram}
\begin{aligned}
\mathbf{G} = \mathbf{A}^T \mathbf{A} = 
\begin{bmatrix}
\mathbf{a}^T_1 \\
\mathbf{a}^T_2 \\
\vdots \\
\mathbf{a}^T_N
\end{bmatrix}
\begin{bmatrix}
\mathbf{a}_1 & \mathbf{a}_2 & \cdots & \mathbf{a}_N
\end{bmatrix},
\end{aligned}
\end{equation}where $\mathbf{G}_1$ and $\mathbf{G}_2$ are the Gram matrices with $g_{1,ij}=g_1(\mathbf{a}_i,\mathbf{a}_j)$, $g_{2,ij}=g_2(\mathbf{b}_i,\mathbf{b}_j)$. The Gram matrix has zero mean in the feature space and is centered at $h_{ij}=\delta_{ij}-1/N$. We refer the reader to \cite{gretton2005measuring, Cao2015DiversityinducedMS} for more details about HSIC.
\end{myDef}

According to the definition of HSIC, the variables are a series of independent observations, a Hilbert space, and another Hilbert space. To facilitate the comparison of the two independent observations, we use ${\bf{Z}}_i$ and ${\bf{Z}}_j$ to represent the self-representation matrix of the $i$-th and the $j$-th view, respectively. Consider a series of $n$ independent observations drawn from $p_\mathbf{ab}$, $\mathbf{Y} \subseteq \bf{Z}_i \times \bf{Z}_j$.
Based on HSIC, the diversity regularizer is defined as
\begin{equation}\label{Divregu}
  {R_d}({{\bf{Z}}_i},{{\bf{Z}}_j}) = \sum\limits_{ij} {HSIC({{\bf{Y}}},{{\bf{Z}}_i},{{\bf{Z}}_j})}.
\end{equation}

\begin{figure}[!htbp]
	\centering
	\includegraphics[width=1\linewidth]{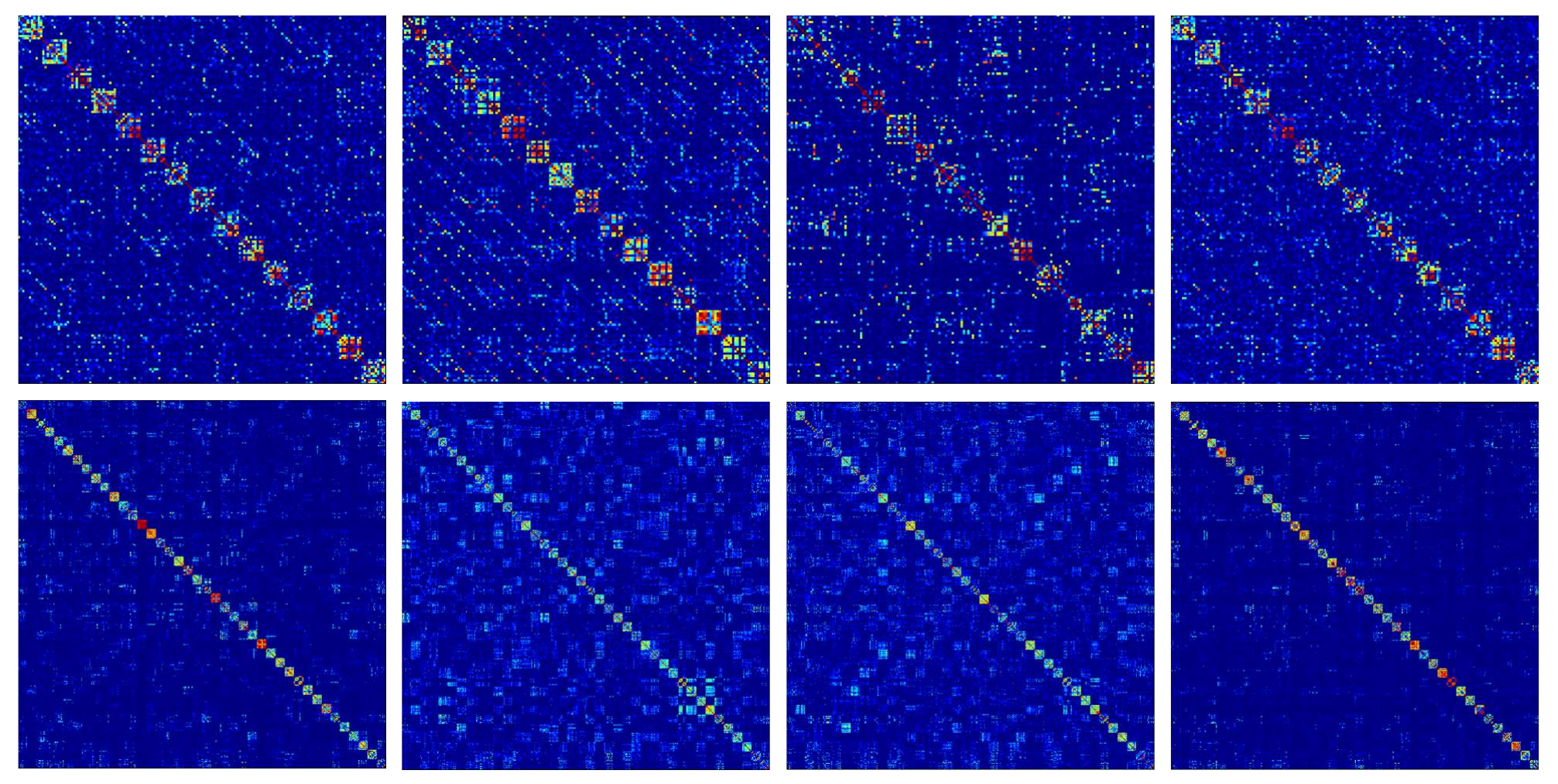}
	\caption{Visualization of the affinity matrices of different views.
             The first three columns are the affinity matrices of view 1, view 2, and view 3 learned by the DSCN \cite{ji2017deep}.
             The last column is the affinity matrix obtained by the MvDSCN for all views.
             The top row is the result on the Yale dataset, and the bottom row is the result on the ORL dataset.}
	\label{affinity}
\end{figure}

The diversity regularizer in Eq. (\ref{Divregu}) can exploit the complementary information from multiple views.
By considering multi-view relations, the objective function is rewritten as

\begin{equation}\label{objective}
 \resizebox{.9\hsize}{!}{$
 \begin{split}
L_{all} = {L_{r}}({{\bf{X}}_i}, {\bf{\hat X}}_i^s, {\bf{\hat X}}_i^c) + {\lambda _1}{L_{s}}({\bf{F}}_i^s, {\bf{F}}_i^c, {{\bf{Z}}_i}, {\bf{Z}}) + {\lambda _2}R_{l_p}({{\bf{Z}}},{{\bf{Z}}_i}) +
\\
{\lambda _3}{R_u}({{\bf{Z}}},{{\bf{Z}}_i}) + {\lambda _4}{R_d}({{\bf{Z}}_i},{{\bf{Z}}_j}),
 \end{split}$}
\end{equation}
where $\lambda_1$, $\lambda_2$, $\lambda_3$, and $\lambda_4$ are scaling factors. Here, we can also consider other types of regularizer, \eg, the nuclear norm \cite{Guangcan2013Robust} or the block diagonal regularizer \cite{lu2019subspace}.
Fig. \ref{affinity} shows the affinity matrix of each view learned independently by the deep subspace clustering network in \cite{ji2017deep} along with the one learned by the MvDSCN on multi-view data. It can be observed that the affinity matrix learned by the MvDSCN has better a block diagonal property and less noise.

\subsection{Optimization}
We use a gradient descent method to solve the problem in Eq. (\ref{objective}).
For the backpropagation (BP) process, the gradients should be derived for each variable.
The encoders and decoders can be updated by standard BP.
Here, we focus on updating the self-representation layer.
Because the optimization problems in Eq. (\ref{objective}) with respect to the view-specific self-representation matrices ${\bf{Z}}_1, {\bf{Z}}_2,...,{\bf{Z}}_v$ and view-consistent matrix ${\bf{Z}}$ are convex, we can obtain the gradients easily. The objective function is represented by $L$. To present the derivation more clearly, we describe $l_p$-norm regularization, which is expressed as follows:
\begin{equation}\label{lp_norm}
\begin{aligned}
\left\| {\bf{A}} \right\|_p = \sqrt[p]{\left( \sum_{i=1}^m \sum_{j=1}^n{|a_{ij}|^p}\right)},
\end{aligned}
\end{equation}
where ${\bf{A}}_{m\times n}$ is any matrix, and its elements are $a_{i j}$. We define $ \sum\limits_{i=1}^m \sum\limits_{j=1}^n{|a_{i j}|^p}$ as $a$. The gradient of $l_p$-norm regularization is as follows:
\begin{equation}\label{grad_A}
\begin{aligned}
\frac{\partial{\left\| {\bf{A}} \right\|_p}}{\partial{a_{ij}}}
& = \frac{1}{p} a^{\frac{1}{p}-1}\frac{\partial{a}}{\partial{a_{ij}}}
\\
& =\frac{1}{p} a^{\frac{1}{p}-1} p |a_{ij}|^{\frac{1}{p}({1-p})} \frac{\partial{a_{ij}}}{\partial{a_{ij}}}
\\
& =(\frac{a_{ij}}{\left\| {\bf{A}} \right\|_p})^{p-1}. \\
\end{aligned}
\end{equation}
Similarly, the gradients of the $l_p$-norm regularizer with respect to ${\bf{Z}}_i$ and ${\bf{Z}}$ are as follows:
\begin{equation}\label{grad_Zi_Zi}
\begin{aligned}
\frac{\partial{\left\| {{\bf{Z}}_i} \right\|_p}}{\partial{{\bf{Z}}_i}}
& =\frac{1}{\left\| {{\bf{Z}}_i} \right\|_{p}^{p-1}} 
\begin{bmatrix}
   {z'_{11}}^{p-1} & {z'_{12}}^{p-1} &\cdots & {z'_{1n}}^{p-1} \\
   \vdots & \vdots & \vdots & \vdots \\
   {z'_{m1}}^{p-1} & {z'_{m2}}^{p-1} &\cdots & {z'_{m n}}^{p-1}
  \end{bmatrix}\\
& =\frac{\bf{\widetilde{Z'}}}{\left\| {{\bf{Z}}_i} \right\|_{p}^{p-1}}, \\
\end{aligned}
\end{equation}

\begin{equation}\label{grad_Z_Z}
\begin{aligned}
\frac{\partial{\left\| {{\bf{Z}}} \right\|_p}}{\partial{{\bf{Z}}}}
& =\frac{1}{\left\| {{\bf{Z}}} \right\|_{p}^{p-1}} 
\begin{bmatrix}
   z_{11}^{p-1} & z_{12}^{p-1} &\cdots & z_{1n}^{p-1} \\
   \vdots & \vdots & \vdots & \vdots \\
   z_{m1}^{p-1} & z_{m2}^{p-1} &\cdots & z_{mn}^{p-1}
  \end{bmatrix}\\
  & =\frac{\bf{\widetilde{Z}}}{\left\| {{\bf{Z}}} \right\|_{p}^{p-1}}. \\
\end{aligned}
\end{equation}
To express equations more concisely, we also introduce $\bf{\widetilde{Z'}}$ and $\bf{\widetilde{Z}}$.
The gradient for ${\bf{Z}}_i$ is
\begin{equation}\label{grad_Zi}
\begin{aligned}
\frac{\partial{L}}{\partial{{\bf{Z}}_i}}
& = 2 \lambda_1 {{{\bf{F}}_i^s}^T {{\bf{F}}_i^s} ({\bf{Z}}_i - \bf{I})}  -
2 \lambda_3 ({{\bf{Z}} - {{\bf{Z}}_i}})  \\
& + 2 \lambda_2 {\frac{\bf{\widetilde{Z'}}}{\left\| {{\bf{Z}}_i} \right\|_{p}^{p-1}}} + \lambda_4 (N-1)^{-2} \sum\limits_{j > i}{({\bf{H}}{{\bf{Z}}_j}{\bf{H}})}^T, \\
\end{aligned}
\end{equation}
and the gradient for ${\bf{Z}}$ is
\begin{equation}\label{grad_Z}
\begin{aligned}
\frac{\partial{L}}{\partial{\bf{Z}}}
= \sum\limits_{i = 1}^v {2 \lambda_1 {{{\bf{F}}_i^c}^T {{\bf{F}}_i^c} ({\bf{Z}} - \bf{I})} + {2 \lambda_2 \frac{\bf{\widetilde{Z}}}{\left\| {{\bf{Z}}} \right\|_{p}^{p-1}}} + 2 \lambda_3 ({\bf{Z}} - {\bf{Z}}_i)}.
\end{aligned}
\end{equation}

We first pre-train the deep autoencoder without the self-representation layer on all multi-view data because the network is difficult to train directly from scratch. This approach also avoids the trivial all-zero solution when minimizing the loss function.
We then use the pre-trained parameters to initialize the convolutional encoder-decoder layers of both the Dnet and Unet.
In the fine-tuning stage, we build a big batch using all the data to minimize the loss function.
The model is trained using Adam \cite{Kingma2015AdamAM} and an initial learning rate of $0.001$.
For the regularization hyper-parameters of the self-representation loss and $l_p$-norm regularizer, we set $\lambda_1 = 1.0 \times 10^{\frac{k}{10} - 3}$, where $k$ is the number of subspaces, $\lambda_2 = 1.0$, $\lambda_3 = 0.1$, and $\lambda_4 = 0.1$.

Our network jointly updates the Dnet and Unet.
Once the network converges, we can use the parameters of the common self-representation layer for all views to construct an affinity matrix $\left( {\left| {\bf{Z}} \right| + {{\left| {\bf{Z}} \right|}^T}} \right)/2$ for spectral clustering.
Similar to \cite{ji2017deep}, our training strategy is unsupervised because we have no access to the labels.
The optimization of the MvDSCN is summarized in Algorithm \ref{alg}.
\begin{algorithm}[!htbp]
	\caption{MvDSCN Optimization}
	\KwIn{Unlabeled multi-view data $\{{\bf{X}}_1,...,{\bf{X}}_v\}$, hyper-parameters $\lambda_1$, $\lambda_2$, $\lambda_3$, and $\lambda_4$, pre-trained epochs $n$, learning rate $\alpha_t$, and initialize parameters $\theta_{\text{Dnet}}$ and $\theta_{\text{Unet}}$ with random values\;
	}
	\For{$j=1$ to $n$}
	{
		Update the autoencoders of $\theta_{\text{Dnet}}$\;
		Update the autoencoders of $\theta_{\text{Unet}}$\;
	}
	\While{not converged}
	{
		Compute the gradient of Eq. \eqref{objective} and update the autoencoders of $\theta_{\text{Dnet}}$ and $\theta_{\text{Unet}}$\;
		Optimize ${{\bf{Z}}_1},{{\bf{Z}}_2},...,{{\bf{Z}}_v}$, and ${\bf{Z}}$ by Eq. \eqref{grad_Zi} and \eqref{grad_Z}\;
	}
	Conduct spectral clustering using affinity matrix ${\bf{Z}}$\;	
	\KwOut{Clustering result $\bf{C}$.}
	\label{alg}
\end{algorithm}

\subsection{Multi-backbone Learning}
\begin{figure}[!htbp]
	\centering
	\includegraphics[width=1\linewidth]{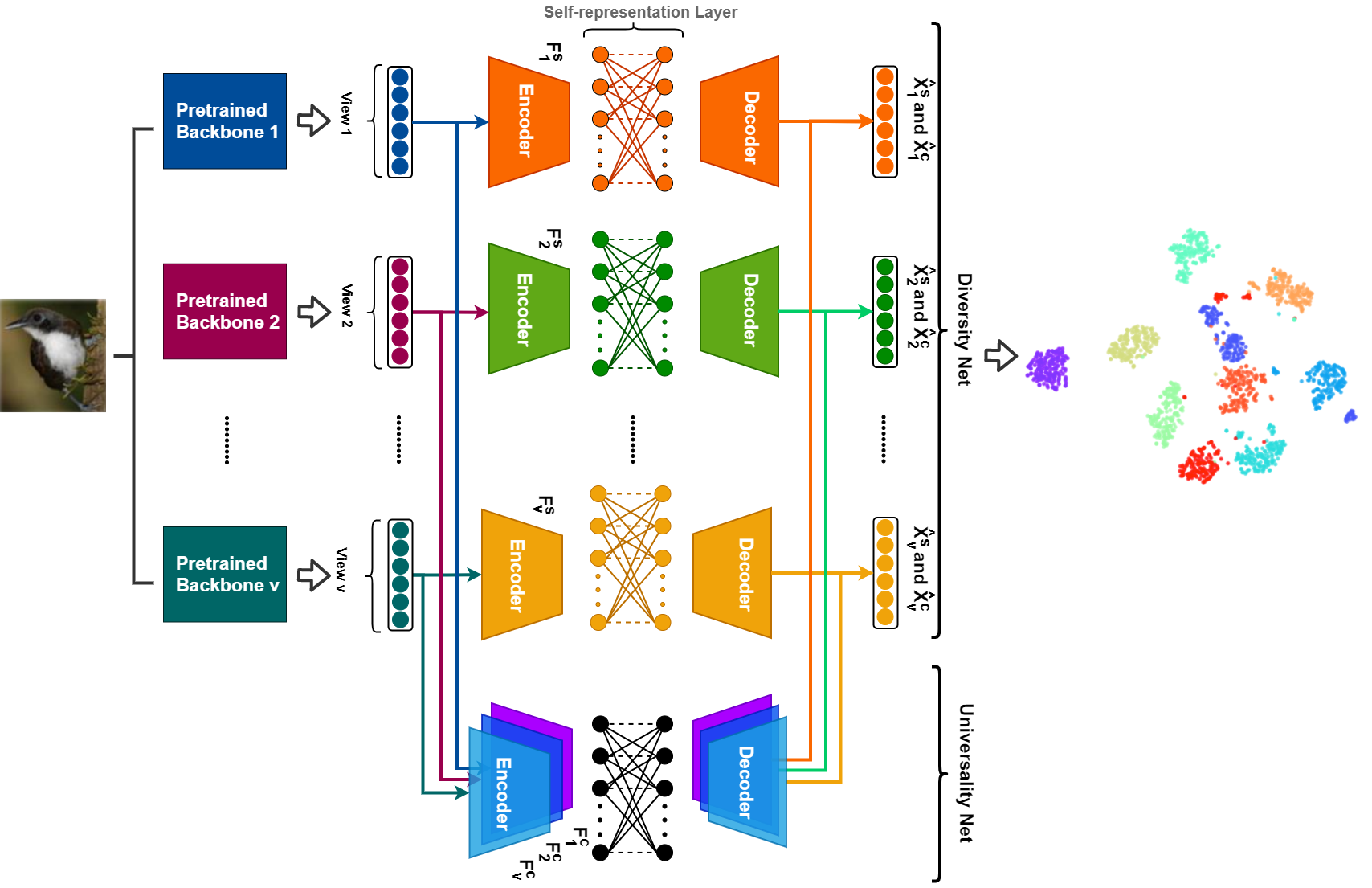}
	\caption{MvDSCN with different pre-trained backbone models.}
	\label{fig:multi-backbone}
\end{figure}

Similar to the range of handcrafted features available, \eg, Gabor, SIFT (scale-invariant feature transform), and LBP (local binary pattern), different kinds of backbones can be used for deep neural networks, such as VGG or ResNet.
Choosing the correct backbone model for different tasks is very challenging.
Inspired by ensemble learning models, \eg, random forest \cite{liaw2002classification} and rotation forest \cite{rodriguez2006rotation}, we fuse different backbones into a unified model, as shown in Fig. \ref{fig:multi-backbone}.
Different types of feature representations are each extracted by the pre-trained VGG and ResNet, and used as the input of the autoencoders.
The parameters of the autoencoders of both $\theta_{Dnet}$ and $\theta_{Unet}$ are initialized by the pre-trained backbones.
The network parameters are then updated using BP.
In this way, the MvDSCN jointly learns multi-feature representations and self-representation coefficients, avoiding model selection.

\subsection{Discussion}

{
\textbf{Connection with latent representation-based methods}}.
Recent works on multi-view subspace clustering focus on learning a latent representation across views using dictionary learning or matrix factorization \cite{Zhao2017MultiViewCV,Zhang2017LatentMS,ZhangGeneralized}.
As shown in Eq. (\ref{latent}), a latent representation $\bf{C}$ is learned for $\bf{X}$, and self-representation is performed on $\bf{C}$.
All views share the same latent representation \cite{Zhang2017LatentMS,ZhangGeneralized} but have a view-specific $\bf{D}$.
\begin{equation}\label{latent}
  {\mathop {\min }\limits_{\{ {\bf{C}},{\bf{D}},{\bf{Z}}\} } \left\| {{\bf{X}} - {\bf{CD}}} \right\|_F^2 + \left\| {{\bf{C}} - {\bf{CZ}}} \right\|_F^2}.
\end{equation}
Like latent representation-based methods, the MvDSCN also learns a hidden representation $\bf{F}$ using autoencoder $\phi$.
An autoencoder can be considered a mapping function that projects the input to a latent space.
The MvDSCN has the following two advantages:
1) Compared with the existing shallow models, the hidden representation $\bf{F}$ is more informative when using deep convolutional autoencoders.
2) The MvDSCN combines feature learning and self-representation in an end-to-end manner. Thus, the multi-view relations can guide both affinity matrix learning and feature learning. Hence, the MvDSCN can learn a suitable affinity matrix and boost the performance of multi-view subspace clustering. 

\textbf{Extension to large-scale tasks}.
We conducted experiments on several small datasets such as Yale and ORL (several hundred samples in total, dozens for each category). We found that a sample can be represented by a linear combination of a set of samples. Many commonly used datasets are larger than Yale and ORL, so in our experimental setup, a sample can be empirically represented by a linear combination of a set of samples. The proposed method is mainly composed of the encoder, self-representation based subspace clustering, and decoder. For the time complexity of the encoder, if the number of convolutional layers is $D$, then the time complexity can be computed as $O\left(\sum\limits_{l=1}^D M_l^2 \cdot K_l^2 \cdot C_{l-1} \cdot C_l \right)$, where $l$ represents the $l$-th convolutional layer, $C$ represents the number of convolutional kernels at this layer, and $K$ represents the size of each convolutional kernel. Since the number of transposed convolutional layers of the decoder is the same as that of the encoder, the time complexity of the decoder can also be computed as $O\left(\sum\limits_{l=1}^D M_l^2 \cdot K_l^2 \cdot C_{l-1} \cdot C_l \right)$. For the time complexity of self-representation based subspace clustering algorithm, if the number of samples is $n$, then the time complexity is generally $O(n^3)$ in that it involves the matrix operations (\ie, SVD decomposition and matrix inversion). Considering that there are $V$ views as inputs, the time complexity of our proposed method is $O\left(2\sum\limits_{
v=1}^V \left(n_v^3+2 \sum\limits_{l=1}^D M_{l v}^2 \cdot K_{l v}^2 \cdot C_{l-1, v} \cdot C_{l v} \right)\right)$.
Hence, we cannot ignore the fact that the time complexity is $O(n^3)$ for the self-representation based subspace clustering algorithms.
To make subspace clustering scalable to large-scale clustering tasks, Peng \etal propose the SSSC (scalable sparse subspace clustering) that adopts a "sampling, clustering, coding, and classifying" strategy to handle both scalability and
the out-of-sample problem \cite{peng2013scalable}.
Although SSSC can cluster an arbitrary number of new samples, the out-of-samples should have a distribution that is similar to the distribution of the training data sampled for subspace clustering.
For the MvDSCN, we can first train a network using the whole batch to obtain clustering result $C$ and then adopt the strategy of SSSC to cluster arbitrary out-of-sample data.
Subspaces $\{{\bf{B}}_1, {\bf{B}}_2,...{\bf{B}}_k\}$ are generated using the clustering result $C$ learned from the MvDSCN.
The representation residual $r_i$ corresponding to each subspace in Eq. (\ref{residual}) is used to obtain the clustering assignment of a new sample as follows:
\begin{equation}\label{residual}
  {r_i} = \frac{{\left\| {{\bf{x}} - {{\bf{B}}_i}{{\bf{a}}_i}} \right\|_2^2}}{{\left\| {{{\bf{a}}_i}} \right\|_2^2}},
\end{equation}
where ${{{\bf{a}}_i}}$ can be obtained by solving a least squares problem.
Alternatively, the nearest neighbor classifier can be used to obtain the cluster assignment of a new sample.
The pseudo labels of a seed dataset are obtained by training a clustering network, and then the pseudo label of the nearest neighbor can be assigned to a new sample.

\section{Experiments}
\label{s4}
In this section, we report the results of extensive experiments conducted to verify the effectiveness of the proposed clustering model. Source code is available at https://github.com/yxjdarren/MvDSCN.

\subsection{Experiment Setup}
\textbf{Datasets.}
We extensively evaluated the multi-view clustering performance of the proposed model on several benchmark multi-view datasets.
\begin{itemize}
  \item  {\textbf{Yale}}\footnote{http://cvc.yale.edu/projects/yalefaces/yalefaces.html} is a widely used face dataset that contains $165$ grayscale images, which are composed of $15$ individuals with $11$ images per person.
Variations of the data include ``center light'', ``with glasses'', ``happy'', ``left light'', ``without glasses'', ``normal'', ``right light'', ``sad'', ``sleepy'', ``surprised'', and ``winking''.
  \item {\textbf{ORL}}\footnote{https://www.cl.cam.ac.uk/research/dtg/attarchive/facedatabase.html} contains $10$ different images of $40$ distinct subjects.
For each subject, the images were taken under varying lighting conditions with different facial expressions (``open/closed eyes'', ``smiling/not smiling'') and facial details (``glasses/no glasses'').
For the face dataset (Yale and ORL), we adjusted the image size to $48\times48$ and extracted three types of features, \ie, intensity features ($4,096$ dimensions), LBP features ($3,304$ dimensions), and Gabor features ($6,750$ dimensions).
The standard LBP features were then extracted from the $72\times80$ loosely cropped image with a histogram size of $59$ over $910$ pixels.
The Gabor feature was dominated by four directions $\theta = {0^\circ, 45^\circ, 90^\circ, 135^\circ }$ and was extracted at a scale of $\lambda = 4$. It had a resolution of $25 \times 30$ pixels and a loose face cropping.
Note that all descriptors except for the intensity features were scaled to have a unit norm.
  \item {\textbf{Still DB}} \cite{still_db} consists of $467$ images with six classes of actions.
The SIFT BoW (bag of words), color SIFT BoW, and shape context BoW features are extracted.
  \item {\textbf{BBCSport}}\footnote{http://mlg.ucd.ie/datasets/bbc.html} contains $544$ documents from the BBCSport website of sports news articles, which are related to two viewpoints in five topical areas and are published in 2004 and 2005. For each sample, there are $3,183$ features for the first view and $3,203$ features for the second view.
      {
   \item {\textbf{ImageNet}} \cite{imagenet_bib} is a widely used large-scale image dataset. ISLVRC 2012 is a subset of ImageNet that consists of 1.3 million samples in the training set and 50,000 samples in the validation set.
   We first trained VGG16 and ResNet50 on the training set and then extracted 2,048 features and 4,096 features on the test set using the pre-trained VGG16 and ResNet50 models, respectively.
   A subset with 10,000 samples was selected from the validation set to evaluate the clustering performance.
   }
\end{itemize}

In addition to multi-feature subspace clustering, the MvDSCN can be easily extended to multi-modal learning by replacing the input with data with different modalities.
We evaluated the proposed deep multi-view subspace clustering methods on the real-world RGB-D Object dataset \cite{Lai2011ALH}.

\begin{figure}[!htbp]
	\centering
	\includegraphics[width=1\linewidth]{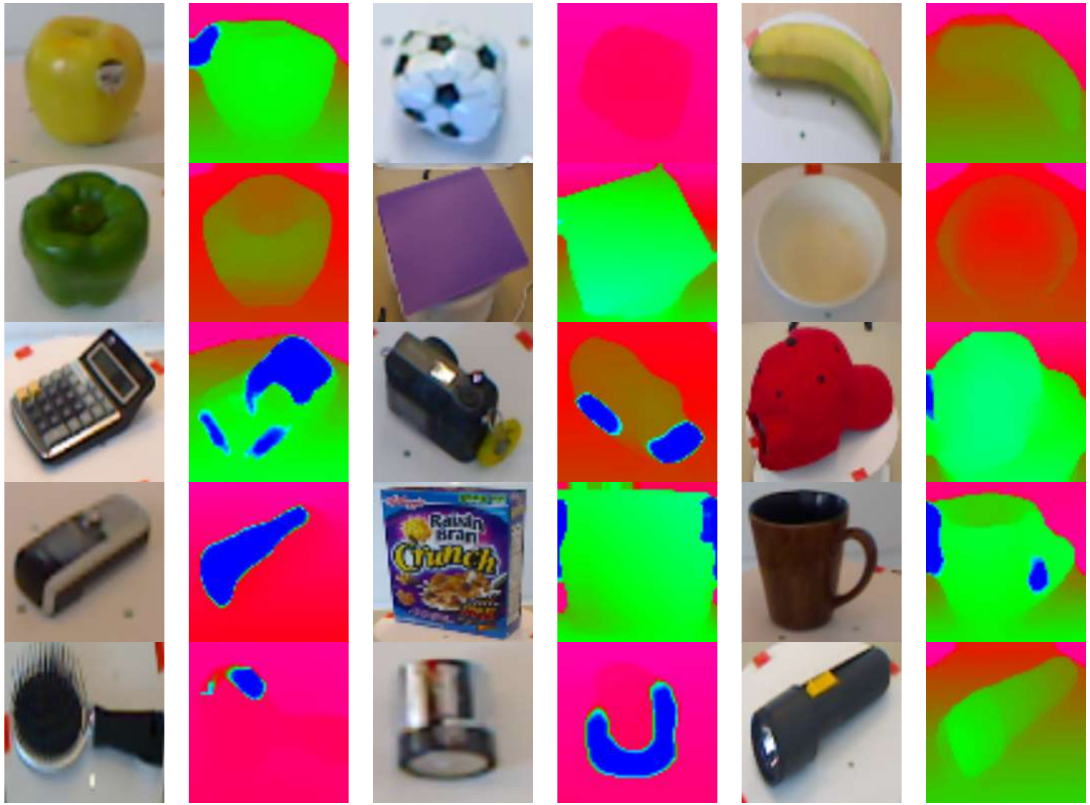}
	\caption{Samples from RGB-D Object dataset. RGB image (left) and the corresponding depth image using a recursive median filter (right).}
	\label{fig:rgbdexample}
\end{figure}

To evaluate the performance of the clustering method on a large-scale dataset, we used the CUB (Caltech-UCSD Birds-200-2011) dataset \cite{WahCUB_200_2011}.
\begin{itemize}
\item  {\textbf{RGB-D Object}} \cite{Lai2011ALH} contains visual and depth images of 300 physically distinct objects taken from multiple views. The objects are organized into $51$ categories arranged by WordNet hypernym--hyponym relationships (similar to ImageNet).
Our experimental datasets were composed of $50$ categories randomly selected from the RGB-D Object dataset, with each class containing $10$ examples.
All visual images and depth images were resized to $64\times64$ pixels.
We applied the median filter recursively until all missing values were filled for visualization.
A subset of the RGB and depth images are shown in Fig. \ref{fig:rgbdexample}.
{
\item  {\textbf{CUB}} \cite{WahCUB_200_2011} contains $200$ categories of bird species with a total of $11,788$ images. In addition to the image view, each category is annotated with $312$ binary attributes (\eg, the attribute group ``belly color'' contains $15$ different color choices).
}
\end{itemize}

\begin{table*}[!htbp]
	\centering
	\caption{Results on four multi-feature datasets (mean $\pm$ standard deviation). Higher values indicate better performance.
	The best values are highlighted in \textcolor{red}{red}, and the second best values are highlighted in \textcolor{blue}{blue}.}
	\begin{tabular}{|c|c|c|c|c|c|}
		\hline
		Datasets & Methods & NMI & ACC & ARI & F-measure\\
		\hline\hline
		\multirow{18}*{Yale}
		& BestSV & 0.654 $\pm$ 0.009 & 0.616 $\pm$ 0.030 & 0.440 $\pm$ 0.011 & 0.475 $\pm$ 0.011 \\
		& LRR  & 0.709 $\pm$ 0.011 & 0.697 $\pm$ 0.000 & 0.515 $\pm$ 0.004 & 0.547 $\pm$ 0.007 \\
		& Min-Disagreement & 0.645 $\pm$ 0.005 & 0.615 $\pm$ 0.043 & 0.433 $\pm$ 0.006 & 0.470 $\pm$ 0.006 \\
		& Co-Reg & 0.648 $\pm$ 0.002 & 0.564 $\pm$ 0.000 & 0.436 $\pm$ 0.002 & 0.466 $\pm$ 0.000 \\
		& RMSC & 0.684 $\pm$ 0.033 & 0.642 $\pm$ 0.036 & 0.485 $\pm$ 0.046 & 0.517 $\pm$ 0.043 \\
		& DSCN & 0.738 $\pm$ 0.006 & 0.727 $\pm$ 0.014 & 0.509 $\pm$ 0.021 & 0.542 $\pm$ 0.019 \\
		& DCSC & 0.744 $\pm$ 0.009 & 0.733 $\pm$ 0.007 & 0.521 $\pm$ 0.011 & 0.556 $\pm$ 0.012 \\
		& DC & 0.756 $\pm$ 0.001 & 0.766 $\pm$ 0.007 & 0.553 $\pm$ 0.017 & 0.579 $\pm$ 0.004 \\
		& LMSC & 0.702 $\pm$ 0.013 & 0.670 $\pm$ 0.012 & 0.472 $\pm$ 0.018 & 0.506 $\pm$ 0.010 \\
		& DMF & 0.782 $\pm$ 0.010 & 0.745 $\pm$ 0.011 & 0.579 $\pm$ 0.002 & 0.601 $\pm$ 0.002 \\
		& FMR & \textcolor{red}{0.832 $\pm$ 0.027} & \textcolor{red}{0.841 $\pm$ 0.010} & \textcolor{red}{0.639 $\pm$ 0.021} & \textcolor{blue}{0.674 $\pm$ 0.044} \\
		& MSCN & 0.769 $\pm$ 0.003 & 0.772 $\pm$ 0.004 & 0.582 $\pm$ 0.012 & 0.598 $\pm$ 0.006 \\
		& DMSC-UDL & 0.785 $\pm$ 0.012 & 0.797 $\pm$ 0.013 & 0.599 $\pm$ 0.016 & 0.647 $\pm$ 0.011 \\
            & CPM & 0.779 $\pm$ 0.026 & 0.781 $\pm$ 0.029 & 0.587 $\pm$ 0.018 & 0.642 $\pm$ 0.023 \\
		& IMVTSC-MVI & 0.796 $\pm$ 0.004 & 0.813 $\pm$ 0.009 & 0.618 $\pm$ 0.003 & \textcolor{red}{0.685 $\pm$ 0.005} \\
        & DCP & 0.793 $\pm$ 0.014 & 0.816 $\pm$ 0.013 & 0.615 $\pm$ 0.016 & 0.673 $\pm$ 0.017 \\
        & SURE & 0.791 $\pm$ 0.013 & 0.821 $\pm$ 0.011 & 0.621 $\pm$ 0.009 & 0.671 $\pm$ 0.013 \\
		& MvDSCN & \textcolor{blue}{0.797 $\pm$ 0.007} & \textcolor{blue}{0.824 $\pm$ 0.004} & \textcolor{blue}{0.626 $\pm$ 0.011} & 0.650 $\pm$ 0.010 \\
		\hline
		\multirow{18}*{ORL}
		& BestSV & 0.903 $\pm$ 0.016 & 0.777 $\pm$ 0.033 & 0.738 $\pm$ 0.001 & 0.711 $\pm$ 0.043 \\
		& LRR  & 0.895 $\pm$ 0.006 & 0.773 $\pm$ 0.003 & 0.724 $\pm$ 0.002 & 0.731 $\pm$ 0.004 \\
		& Min-Disagreement & 0.816 $\pm$ 0.001 & 0.734 $\pm$ 0.040 & 0.621 $\pm$ 0.003 & 0.663 $\pm$ 0.003 \\
		& Co-Reg & 0.853 $\pm$ 0.003 & 0.715 $\pm$ 0.000 & 0.602 $\pm$ 0.004 & 0.615 $\pm$ 0.000 \\
		& RMSC & 0.872 $\pm$ 0.012 & 0.723 $\pm$ 0.025 & 0.645 $\pm$ 0.029 & 0.654 $\pm$ 0.028 \\
		& DSCN & 0.883 $\pm$ 0.005 & 0.801 $\pm$ 0.009 & 0.704 $\pm$ 0.012 & 0.711 $\pm$ 0.011 \\
		& DCSC & 0.893 $\pm$ 0.003 & 0.811 $\pm$ 0.003 & 0.709 $\pm$ 0.021 & 0.718 $\pm$ 0.004 \\
		& DC	& 0.865 $\pm$ 0.011 & 0.788 $\pm$ 0.002 & 0.684 $\pm$ 0.007 & 0.701 $\pm$ 0.008 \\
		& LMSC & 0.931 $\pm$ 0.011 & 0.819 $\pm$ 0.017 & 0.769 $\pm$ 0.044 & 0.758 $\pm$ 0.009 \\
		& DMF & 0.933 $\pm$ 0.010 & 0.823 $\pm$ 0.021 & 0.783 $\pm$ 0.001 & 0.773 $\pm$ 0.002 \\
		& FMR & 0.932 $\pm$ 0.012 & \textcolor{blue}{0.855 $\pm$ 0.014} & \textcolor{blue}{0.803 $\pm$ 0.016} & 0.802 $\pm$ 0.016 \\
		& MSCN & 0.928 $\pm$ 0.001 & 0.833 $\pm$ 0.008 & 0.790 $\pm$ 0.005 & 0.787 $\pm$ 0.001 \\
		& DMSC-UDL & 0.935 $\pm$ 0.012 & 0.838 $\pm$ 0.017 & 0.793 $\pm$ 0.011 & 0.793 $\pm$ 0.013 \\
        & CPM & 0.929 $\pm$ 0.015 & 0.837 $\pm$ 0.017 & 0.778 $\pm$ 0.021 & \textcolor{blue}{0.826 $\pm$ 0.011} \\
		& IMVTSC-MVI & \textcolor{blue}{0.937 $\pm$ 0.010} & 0.849 $\pm$ 0.018 & 0.802 $\pm$ 0.014 & 0.809 $\pm$ 0.018 \\
        & DCP & 0.933 $\pm$ 0.012 & 0.838 $\pm$ 0.016 & 0.786 $\pm$ 0.016 & 0.783 $\pm$ 0.017 \\
        & SURE & 0.925 $\pm$ 0.015 & 0.843 $\pm$ 0.012 & 0.791 $\pm$ 0.011 & 0.788 $\pm$ 0.021 \\
		& MvDSCN & \textcolor{red}{0.943 $\pm$ 0.002} & \textcolor{red}{0.870 $\pm$ 0.006} & \textcolor{red}{0.819 $\pm$ 0.001} & \textcolor{red}{0.834 $\pm$ 0.012} \\
		\hline
		\multirow{18}*{Still DB}
		& BestSV & 0.104 $\pm$ 0.078 & 0.297 $\pm$ 0.089 & 0.063 $\pm$ 0.001 & 0.221 $\pm$ 0.064 \\
		& LRR  & 0.109 $\pm$ 0.030 & 0.306 $\pm$ 0.039 & 0.066 $\pm$ 0.002 & 0.240 $\pm$ 0.052 \\
		& Min-Disagreement & 0.097 $\pm$ 0.005 & 0.336 $\pm$ 0.014 & 0.103 $\pm$ 0.013 & 0.223 $\pm$ 0.004 \\
		& Co-Reg & 0.093 $\pm$ 0.016 & 0.263 $\pm$ 0.024 & 0.092 $\pm$ 0.004 & 0.226 $\pm$ 0.035 \\
		& RMSC & 0.106 $\pm$ 0.056 & 0.285 $\pm$ 0.020 & 0.113 $\pm$ 0.063 & 0.232 $\pm$ 0.021 \\
		& DSCN  & 0.216 $\pm$ 0.011 & 0.323 $\pm$ 0.006 & 0.145 $\pm$ 0.002 & 0.293 $\pm$ 0.019 \\
		& DCSC & 0.222 $\pm$ 0.008 & 0.325 $\pm$ 0.007 & 0.148 $\pm$ 0.003 & 0.301 $\pm$ 0.002 \\
		& DC  & 0.199 $\pm$ 0.003 & 0.315 $\pm$ 0.001 & 0.131 $\pm$ 0.001 & 0.280 $\pm$ 0.011 \\
		& LMSC & 0.137 $\pm$ 0.032 & 0.328 $\pm$ 0.029 & 0.088 $\pm$ 0.007 & 0.269 $\pm$ 0.055 \\
		& DMF & 0.154 $\pm$ 0.010 & 0.336 $\pm$ 0.017 & 0.124 $\pm$ 0.001 & 0.265 $\pm$ 0.005 \\
		& FMR & 0.230 $\pm$ 0.005 & 0.344 $\pm$ 0.009 & 0.130 $\pm$ 0.012 & 0.305 $\pm$ 0.006 \\
		& MSCN & 0.168 $\pm$ 0.001 & 0.312 $\pm$ 0.008 & 0.133 $\pm$ 0.005 & 0.261 $\pm$ 0.001 \\
		& DMSC-UDL & \textcolor{red}{0.249 $\pm$ 0.012} & \textcolor{blue}{0.369 $\pm$ 0.013} & \textcolor{blue}{0.155 $\pm$ 0.011} & \textcolor{red}{0.326 $\pm$ 0.012} \\
        & CPM & 0.201 $\pm$ 0.008 & 0.347 $\pm$ 0.012 & 0.114 $\pm$ 0.009 & 0.292 $\pm$ 0.014 \\
		& IMVTSC-MVI & 0.229 $\pm$ 0.006 & 0.362 $\pm$ 0.008 & 0.140 $\pm$ 0.007 & 0.294 $\pm$ 0.010 \\
        & DCP & 0.232 $\pm$ 0.011 & 0.365 $\pm$ 0.016 & 0.146 $\pm$ 0.012 & 0.301 $\pm$ 0.016 \\
        & SURE & 0.227 $\pm$ 0.014 & 0.363 $\pm$ 0.015 & 0.142 $\pm$ 0.009 & 0.296 $\pm$ 0.012 \\
		& MvDSCN & \textcolor{blue}{0.245 $\pm$ 0.020} & \textcolor{red}{0.377 $\pm$ 0.023} & \textcolor{red}{0.169 $\pm$ 0.003} & \textcolor{blue}{0.320 $\pm$ 0.015} \\
		\hline
		\multirow{18}*{BBCSport}
		& BestSV & 0.715 $\pm$ 0.060 & 0.836 $\pm$ 0.037 & 0.659 $\pm$ 0.005 & 0.768 $\pm$ 0.038 \\
		& LRR  & 0.690 $\pm$ 0.019 & 0.832 $\pm$ 0.026 & 0.667 $\pm$ 0.008 & 0.774 $\pm$ 0.023 \\
		& Min-Disagreement & 0.776 $\pm$ 0.019 & 0.797 $\pm$ 0.049 & 0.783 $\pm$ 0.034 & 0.260 $\pm$ 0.013 \\
		& Co-Reg & 0.718 $\pm$ 0.003 & 0.564 $\pm$ 0.000 & 0.696 $\pm$ 0.001 & 0.766 $\pm$ 0.002 \\
		& RMSC & 0.608 $\pm$ 0.007 & 0.737 $\pm$ 0.003 & 0.723 $\pm$ 0.025 & 0.655 $\pm$ 0.002 \\
		& DSCN  & 0.652 $\pm$ 0.000 & 0.821 $\pm$ 0.000 &  0.856 $\pm$ 0.001 & 0.683 $\pm$ 0.001 \\
		& DCSC & 0.683 $\pm$ 0.001 & 0.843 $\pm$ 0.000 & 0.864 $\pm$ 0.012 & 0.712 $\pm$ 0.002 \\
		& DC	& 0.556 $\pm$ 0.001 & 0.724 $\pm$ 0.000 & 0.781 $\pm$ 0.000 & 0.492 $\pm$ 0.000 \\
		& LMSC & 0.826 $\pm$ 0.006 & 0.900 $\pm$ 0.044 & 0.893 $\pm$ 0.012 & 0.887 $\pm$ 0.071 \\
		& DMF & 0.821 $\pm$ 0.003 & 0.890 $\pm$ 0.031 & 0.883 $\pm$ 0.012 & \textcolor{blue}{0.889 $\pm$ 0.001} \\
		& FMR & \textcolor{red}{0.851 $\pm$ 0.018} & \textcolor{red}{0.948 $\pm$ 0.010} & \textcolor{red}{0.913 $\pm$ 0.014} & 0.883 $\pm$ 0.009 \\
		& MSCN & 0.813 $\pm$ 0.002 & 0.888 $\pm$ 0.003 & 0.859 $\pm$ 0.001 & 0.854 $\pm$ 0.002 \\
		& DMSC-UDL & 0.823 $\pm$ 0.006 & 0.895 $\pm$ 0.004 & 0.862 $\pm$ 0.007 & 0.857 $\pm$ 0.005 \\
        & CPM & 0.802 $\pm$ 0.028 & 0.891 $\pm$ 0.019 & 0.856 $\pm$ 0.031 & 0.834 $\pm$ 0.025 \\
		& IMVTSC-MVI & 0.833 $\pm$ 0.003 & 0.930 $\pm$ 0.002 & 0.889 $\pm$ 0.005 & \textcolor{red}{0.894 $\pm$ 0.003} \\
        & DCP & 0.828 $\pm$ 0.009 & 0.923 $\pm$ 0.013 & 0.886 $\pm$ 0.008 & 0.882 $\pm$ 0.011 \\
        & SURE & 0.824 $\pm$ 0.011 & 0.927 $\pm$ 0.008 & 0.892 $\pm$ 0.013 & 0.887 $\pm$ 0.012 \\
		& MvDSCN & \textcolor{blue}{0.835 $\pm$ 0.000} & \textcolor{blue}{0.931 $\pm$ 0.001} & \textcolor{blue}{0.909 $\pm$ 0.001} & 0.860 $\pm$ 0.000 \\
		\hline
	\end{tabular}
	\label{table:multiview}
\end{table*}

\begin{table*}[!htbp]
	\centering
	\caption{Results on the ImageNet test dataset (mean $\pm$ standard deviation). Higher values indicate better performance.}
	\begin{tabular}{|c|c|c|c|c|c|}
		\hline
		Datasets & Methods & NMI & ACC & ARI & F-measure\\
		\hline\hline
		\multirow{3}*{ImageNet}
		& VGG16 & 0.724 $\pm$ 0.002 & 0.569 $\pm$ 0.003 & 0.457 $\pm$ 0.002 & 0.463 $\pm$ 0.004 \\
		& ResNet50  & 0.733 $\pm$ 0.002 & 0.593 $\pm$ 0.002 & 0.471 $\pm$ 0.002 & 0.476 $\pm$ 0.005 \\
		& MvDSCN  & \textbf{0.749 $\pm$ 0.004} & \textbf{0.610 $\pm$ 0.002} & \textbf{0.502 $\pm$ 0.007} & \textbf{0.506 $\pm$ 0.003} \\
		\hline
	\end{tabular}
	\label{table:ImageNet}
\end{table*}

\textbf{Comparison methods.}
The performance of the MvDSCN is compared with SOTA subspace clustering methods, consisting of shallow and deep single-view clustering algorithms and multi-view clustering algorithms.
For all single-view clustering algorithms, the performance of the best view is reported.
\begin{itemize}
  \item {\textbf{BestSV}} \cite{Ng2001OnSC} reports the result of the individual view that achieves the best spectral clustering performance with a single view of data.
  \item {\textbf{LRR}} \cite{Guangcan2013Robust} seeks the lowest-rank representation among all the candidates that can represent the data samples as linear combinations of the bases in a given dictionary with the best single view.
  \item {\textbf{RMSC}} \cite{Xia2014RobustMS} is a method that uses low-rank and sparse decomposition.
It first constructs a transition probability matrix from every single view and then uses these matrices to recover a shared low-rank transition probability matrix as a primary input to the standard Markov chain method for clustering.
  \item {\textbf{DSCN}} \cite{Peng2016DeepSC} is a deep autoencoder framework for subspace clustering with the best single view. 
  \item {\textbf{DCSC}} \cite{Jiang2018WhenTL} imposes a self-paced regularizer on the loss, which yields a robust deep subspace clustering algorithm.
  \item {\textbf{DC}} \cite{Caron2018DeepCF} is a scalable clustering approach for the unsupervised learning of convolutional networks.
It iterates between clustering with k-means and updating its weights by predicting the cluster assignments as pseudo-labels in a discriminative loss.
\end{itemize}

\begin{itemize}
  \item {\textbf{Min-Disagreement}} \cite{Sa2005SpectralCW} creates a bipartite graph and is based on the concept of minimizing disagreement.
  \item {\textbf{Co-Reg SPC}} \cite{Kumar2011CoregularizedMS} uses spectral clustering objective functions that implicitly combine graphs from multiple views of the data to achieve a better clustering result.
  \item {\textbf{DMF}} \cite{Zhao2017MultiViewCV} learns the hierarchical semantics of multi-view data through semi-nonnegative matrix factorization.
  \item {\textbf{LMSC}} \cite{Zhang2017LatentMS} seeks the underlying latent representation and simultaneously performs data reconstruction based on the learned latent representation.
  {\item{\textbf{FMR}} \cite{Li2019FlexibleMR} explores the complementary information among multiple views and extends the multi-view representation learning to nonlinear cases using HSIC.}
  \item {\textbf{MSCN}} \cite{Abavisani2018DeepMS} observes that spatial fusion methods in a deep multi-modal subspace clustering task rely on spatial correspondences among the modalities.
  \item{\textbf{DMSC-UDL}} \cite{wang2020deep} combines global and local structures with self-expression layer. The global and local structures help each other forward and achieve a small distance between samples of the same cluster.
  \item{\textbf{CPM}} \cite{DPML_CPM} translates the task of learning latent multi-view representation into a degradation process resembling data transmission.
  \item{\textbf{IMVTSC-MVI}} \cite{wen2021unified} jointly introduces the tensor low-rank representation constraint and semantic consistency-based graph constraint.
  \item{\textbf{DCP}} \cite{DCP} provides a novel insight to the community that the cross-view consistency learning and data recovery are with intrinsic connections in the framework of information theory.
  \item{\textbf{SURE}} \cite{SURE} is a novel contrastive learning paradigm which uses the available pairs as positives and randomly chooses some cross-view samples as negatives.
\end{itemize}

\textbf{Evaluation metrics.}
Following the setting in \cite{Zhao2017MultiViewCV,Zhang2017LatentMS},
four popular metrics were used to evaluate the clustering quality, normalized mutual information (\textbf{NMI}), accuracy (\textbf{ACC}), adjusted Rand index (\textbf{ARI}), and \textbf{F-measure}. We optimized all the parameters to achieve the best performance of the comparison method. We ran each method $30$ times and reported the average performance and standard deviation.

\begin{table*}[!htbp]
	\centering
	\caption{Results on the four multi-feature datasets (mean $\pm$ standard deviation). Higher values indicate better performance.}
	\begin{tabular}{|c|c|c|c|c|c|}
		\hline
		Datasets & Views & NMI & ACC & ARI & F-measure\\
		\hline\hline
		\multirow{4}*{Yale}
		& View1 & 0.738 $\pm$ 0.006 & 0.727 $\pm$ 0.014 & 0.509 $\pm$ 0.021 & 0.542 $\pm$ 0.019 \\
		& View2  & 0.613 $\pm$ 0.007 & 0.598 $\pm$ 0.008 & 0.401 $\pm$ 0.008 & 0.439 $\pm$ 0.008 \\
		& View3 & 0.545 $\pm$ 0.009 & 0.522 $\pm$ 0.012 & 0.267 $\pm$ 0.011 & 0.311 $\pm$ 0.011 \\
		& ALL  & \textbf{0.797 $\pm$ 0.007} & \textbf{0.824 $\pm$ 0.004} & \textbf{0.626 $\pm$ 0.011} & \textbf{0.650 $\pm$ 0.010} \\
		\hline
		\multirow{4}*{ORL}
		& View1	& 0.883 $\pm$ 0.005 & 0.801 $\pm$ 0.009 & 0.704 $\pm$ 0.012 & 0.711 $\pm$ 0.011 \\
		& View2  & 0.793 $\pm$ 0.011 & 0.627 $\pm$ 0.024 & 0.504 $\pm$ 0.023 & 0.516 $\pm$ 0.023 \\
		& View3 & 0.764 $\pm$ 0.009 & 0.580 $\pm$ 0.024 & 0.458 $\pm$ 0.024 & 0.471 $\pm$ 0.023 \\
		& ALL & \textbf{0.943 $\pm$ 0.002} & \textbf{0.870 $\pm$ 0.006} & \textbf{0.819 $\pm$ 0.001} & \textbf{0.834 $\pm$ 0.012} \\
		\hline
		\multirow{4}*{Still DB}
		& View1	& 0.113 $\pm$ 0.001 & 0.329 $\pm$ 0.004 & 0.083 $\pm$ 0.003 & 0.243 $\pm$ 0.014 \\
		& View2  & 0.216 $\pm$ 0.011 & 0.323 $\pm$ 0.006 & 0.145 $\pm$ 0.002 & 0.293 $\pm$ 0.019 \\
		& View3 & 0.211 $\pm$ 0.002 & 0.313 $\pm$ 0.012 & 0.142 $\pm$ 0.014 & 0.289 $\pm$ 0.007 \\
		& ALL & \textbf{0.245 $\pm$ 0.020} & \textbf{0.377 $\pm$ 0.023} & \textbf{0.169 $\pm$ 0.003} & \textbf{0.320 $\pm$ 0.015} \\
		\hline
		\multirow{3}*{BBCSport}
		& View1	& 0.617 $\pm$ 0.000 & 0.801 $\pm$ 0.000 & 0.847 $\pm$ 0.001 & 0.559 $\pm$ 0.000 \\
		& View2  & 0.652 $\pm$ 0.000 & 0.821 $\pm$ 0.000 & 0.856 $\pm$ 0.001 & 0.683 $\pm$ 0.001 \\
		& ALL & \textbf{0.835 $\pm$ 0.000} & \textbf{0.931 $\pm$ 0.001} & \textbf{0.909 $\pm$ 0.001} & \textbf{0.860 $\pm$ 0.000} \\
		\hline
	\end{tabular}
	\label{versus}
\end{table*}

\begin{table*}[!htbp]
	\centering
	\caption{Results on the RGB-D Object dataset (mean $\pm$ standard deviation). Higher values indicate better performance.
	The best values are highlighted in \textcolor{red}{red} and the second best values are highlighted in \textcolor{blue}{blue}.}
	\begin{tabular}{|c|c|c|c|c|c|}
		\hline
		Datasets & Methods & NMI & ACC & ARI & F-measure\\
		\hline\hline
		\multirow{18}*{RGB-D Object}
		& BestSV & 0.554 $\pm$ 0.006 & 0.278 $\pm$ 0.001 & 0.106 $\pm$ 0.006 & 0.125 $\pm$ 0.006 \\
		& LRR & 0.589 $\pm$ 0.002 & 0.299 $\pm$ 0.010 & 0.143 $\pm$ 0.002 & 0.156 $\pm$ 0.001 \\
		& Min-Disagreement & 0.605 $\pm$ 0.008 & 0.332 $\pm$ 0.002 & 0.160 $\pm$ 0.013 & 0.177 $\pm$ 0.011 \\
		& Co-Reg & 0.602 $\pm$ 0.007 & 0.268 $\pm$ 0.003 & 0.155 $\pm$ 0.020 & 0.175 $\pm$ 0.018 \\
		& RMSC & 0.603 $\pm$ 0.006 & 0.341 $\pm$ 0.015 & 0.162 $\pm$ 0.010 & 0.178 $\pm$ 0.010 \\
		& DSCN & 0.589 $\pm$ 0.004 & 0.339 $\pm$ 0.006 & 0.163 $\pm$ 0.004 & 0.179 $\pm$ 0.004 \\
		& DCSC & 0.591 $\pm$ 0.002 & 0.340 $\pm$ 0.002 & 0.170 $\pm$ 0.001 & 0.182 $\pm$ 0.003 \\
		& DC & 0.594 $\pm$ 0.003 & 0.343 $\pm$ 0.002 & 0.177 $\pm$ 0.004 & 0.184 $\pm$ 0.004 \\
		& LMSC & 0.593 $\pm$ 0.030 & 0.335 $\pm$ 0.028 & 0.151 $\pm$ 0.035 & 0.167 $\pm$ 0.034 \\
		& DMF & 0.549 $\pm$ 0.004 & 0.286 $\pm$ 0.006 & 0.107 $\pm$ 0.002 & 0.123 $\pm$ 0.001 \\
		& FMR & 0.576 $\pm$ 0.007 & 0.284 $\pm$ 0.003 & 0.144 $\pm$ 0.006 & 0.146 $\pm$ 0.002 \\
		& MSCN & 0.608 $\pm$ 0.001 & 0.354 $\pm$ 0.003 & 0.190 $\pm$ 0.002 & 0.203 $\pm$ 0.004 \\
    	& DMSC-UDL & 0.611 $\pm$ 0.009 & 0.359 $\pm$ 0.012 & 0.197 $\pm$ 0.010 & 0.211 $\pm$ 0.012 \\
        & CPM & 0.606 $\pm$ 0.012 & 0.321 $\pm$ 0.011 & 0.164 $\pm$ 0.015 & 0.209 $\pm$ 0.018 \\
		& IMVTSC-MVI  & 0.620 $\pm$ 0.001 & 0.367 $\pm$ 0.004 & 0.196 $\pm$ 0.003 & 0.223 $\pm$ 0.009 \\
        & DCP  & \textcolor{blue}{0.627 $\pm$ 0.005} & 0.369 $\pm$ 0.011 & 0.199 $\pm$ 0.008 & 0.221 $\pm$ 0.013 \\
        & SURE  & 0.622 $\pm$ 0.008 & \textcolor{blue}{0.373 $\pm$ 0.009} & \textcolor{blue}{0.202 $\pm$ 0.012} & \textcolor{red}{0.228 $\pm$ 0.017} \\
		& MvDSCN & \textcolor{red}{0.639 $\pm$ 0.003} & \textcolor{red}{0.388 $\pm$ 0.005} & \textcolor{red}{0.210 $\pm$ 0.004} & \textcolor{blue}{0.225 $\pm$ 0.004} \\
		\hline
	\end{tabular}
	\label{table:multi-modal}
\end{table*}

\subsection{Results of Multi-feature Subspace Clustering}
The multi-view clustering performance of different methods is presented in Table \ref{table:multiview}.
Our proposed method shows very competitive performance on multi-feature datasets.
For ORL, we raise the performance by approximately $0.6\%$ in NMI, $1.5\%$ in ACC, $1.6\%$ in ARI, and $0.8\%$ in F-measure.
On Still DB, our method gains improvements of approximately $0.8\%$ and $1.4\%$ over the second-best method in terms of ACC and ARI, respectively.
In addition, for the Yale and BBCSport datasets, the performance of the proposed method is the second-best performance, exceeded by FMR. Compared with other datasets, Yale and BBCSport have less information on samples and views. FMR proposes to construct a latent representation by encouraging it to be similar to different views in a weighted way, which implicitly enforces it to encode complementary information from multiple views. Compared with MvDSCN, FMR pays more attention to the information in the view, which can still perform well in small-scale datasets with sparse view information. Since MvDSCN consists of two sub-networks and embeds multi-view relations into feature learning to obtain a better clustering effect, data with relatively large-scale samples and rich multi-view information can better enable MvDSCN to learn. Overall, MvDSCN has more potential in handling large-scale and view-rich data.

Experimental results demonstrate the effectiveness of the MvDSCN on the multi-feature subspace clustering task. The performance improvement is due to two aspects, \ie, the end-to-end manner for learning the affinity matrix, and multi-view relations embedded into both the feature learning and self-representation.

{In contrast to the four datasets (ORL, Still DB, Yale, and BBCSport) that are provided with different types of handcrafted features, we used different pre-trained models, \ie, VGG16 and ResNet50, to obtain two views for ImageNet.
We used VGG16 and ResNet50 as the backbones, in which VGG16 contains 13 convolutional layers and 3 full connection layers, and ResNet50 contains 4 residual blocks. The feature dimensions extracted by VGG16 and ResNet50 are 2,048 and 4,096, respectively, by taking images in the ImageNet with dimensions 224 $\times$ 224 as input.
The models pre-trained on the ImageNet training set can effectively exploit the large-scale dataset in representation learning.
The experimental results are summarized in Table \ref{table:ImageNet}.
These results demonstrate the superior performance of the MvDSCN, which outperforms the deep subspace clustering network (DSCN) \cite{ji2017deep} using only VGG16 features or ResNet50 features}.

\begin{figure}[!htbp]
	\centering
	\includegraphics[width=1\linewidth]{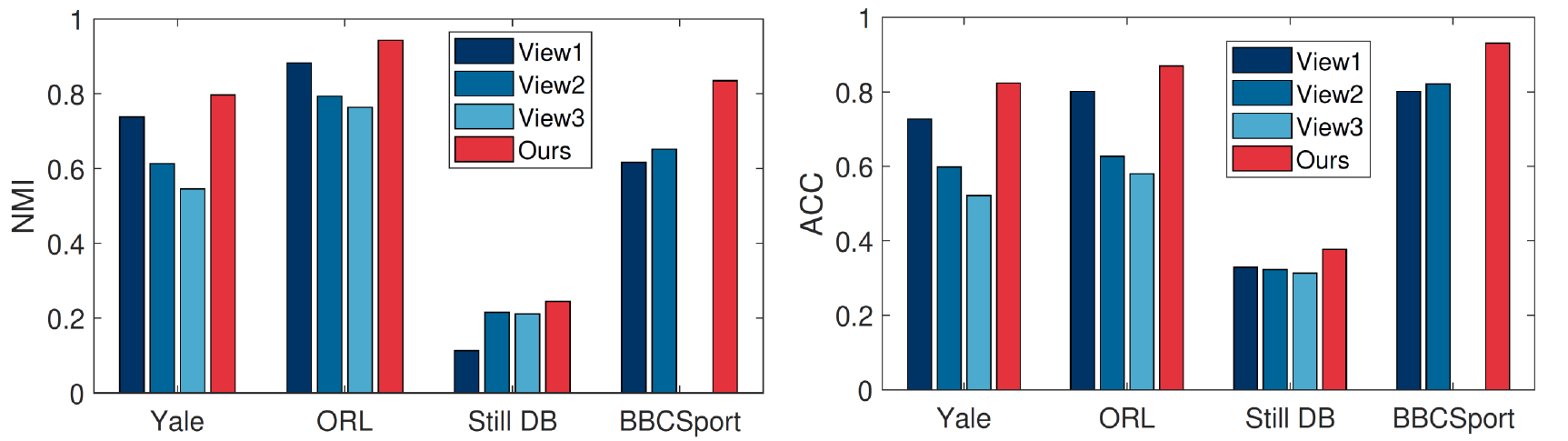}
	\caption{{Performance comparison of the DSCN \cite{ji2017deep} with each view and the MvDSCN versus NMI and ACC.}}
	\label{fig:svsm}
\end{figure}

\begin{table*}[!htbp]
	\centering
	\caption{Results on the CUB dataset (mean $\pm$ standard deviation). Higher values indicate better performance.
	The best values are highlighted in \textcolor{red}{red} and the second best values are highlighted in \textcolor{blue}{blue}.}
	\begin{tabular}{|c|c|c|c|c|c|}
		\hline
		Datasets & Methods & NMI & ACC & ARI & F-measure\\
		\hline\hline
		\multirow{18}*{CUB}
		& BestSV & 0.532 $\pm$ 0.021 & 0.290 $\pm$ 0.017 & 0.158 $\pm$ 0.021 & 0.178 $\pm$ 0.009 \\
		& LRR & 0.644 $\pm$ 0.007 & 0.345 $\pm$ 0.003 & 0.219 $\pm$ 0.006 & 0.194 $\pm$ 0.002 \\
		& Min-Disagreement & - & - & - & - \\
		& Co-Reg & - & - & - & - \\
		& RMSC & - & - & - & - \\
		& DSCN & 0.724 $\pm$ 0.001 & 0.584 $\pm$ 0.002 & 0.563 $\pm$ 0.001 & 0.514 $\pm$ 0.002 \\
		& DCSC & 0.733 $\pm$ 0.002 & 0.599 $\pm$ 0.001 & 0.568 $\pm$ 0.003 & 0.522 $\pm$ 0.002 \\
		& DC & 0.706 $\pm$ 0.004 & 0.550 $\pm$ 0.010 & 0.537 $\pm$ 0.001 & 0.502 $\pm$ 0.002 \\
		& LMSC & 0.792 $\pm$ 0.010 & 0.662 $\pm$ 0.013 & 0.592 $\pm$ 0.009 & 0.596 $\pm$ 0.022 \\
		& DMF & 0.805 $\pm$ 0.002 & 0.711 $\pm$ 0.001 & 0.640 $\pm$ 0.001 & 0.646 $\pm$ 0.003 \\
		& FMR & - & - & - & - \\
		& MSCN & 0.843 $\pm$ 0.003 & 0.772 $\pm$ 0.001 & 0.749 $\pm$ 0.003 & 0.755 $\pm$ 0.005 \\
		& DMSC-UDL & 0.851 $\pm$ 0.009 & 0.775 $\pm$ 0.007 & 0.751 $\pm$ 0.009 & 0.756 $\pm$ 0.006 \\
        & CPM & 0.859 $\pm$ 0.013 & 0.766 $\pm$ 0.009 & 0.739 $\pm$ 0.018 & 0.747 $\pm$ 0.011 \\
		& IMVTSC-MVI & 0.861 $\pm$ 0.006 & \textcolor{red}{0.797 $\pm$ 0.005} & 0.768 $\pm$ 0.003 & \textcolor{red}{0.794 $\pm$ 0.002} \\
        & DCP & \textcolor{blue}{0.872 $\pm$ 0.011} & 0.789 $\pm$ 0.009 & 0.766 $\pm$ 0.007 & 0.784 $\pm$ 0.007 \\
        & SURE & 0.869 $\pm$ 0.008 & 0.792 $\pm$ 0.011 & \textcolor{blue}{0.769 $\pm$ 0.012} & 0.788 $\pm$ 0.015 \\
		& MvDSCN & \textcolor{red}{0.878 $\pm$ 0.004} & \textcolor{blue}{0.795 $\pm$ 0.002} & \textcolor{red}{0.771 $\pm$ 0.002} & \textcolor{blue}{0.792 $\pm$ 0.006} \\
		\hline
	\end{tabular}
	\label{table:cub}
\end{table*}
\textbf{Single view versus multiple views}.
To further investigate the improvements obtained by our method, we compared our method with the DSCN \cite{ji2017deep} on data with only a single view.
Fig. \ref{fig:svsm} shows the detailed results on different datasets.
According to Table \ref{versus}, clustering with multiple views consistently outperforms clustering with each single view, which empirically proves that clustering with multiple views is more robust than that with a single view.

\begin{table*}[!htbp]
	\centering
	\caption{Results on the RGB-D Object dataset (mean $\pm$ standard deviation). Higher values indicate better performance.}
	\begin{tabular}{|c|c|c|c|c|c|}
		\hline
		Datasets & Views & NMI & ACC & ARI & F-measure\\
		\hline\hline
		\multirow{3}*{RGB-D Object}
		& RGB & 0.589 $\pm$ 0.004 & 0.339 $\pm$ 0.006 & 0.163 $\pm$ 0.004 & 0.179 $\pm$ 0.004 \\
		& Depth  & 0.576 $\pm$ 0.004 & 0.300 $\pm$ 0.005 & 0.131 $\pm$ 0.004 & 0.147 $\pm$ 0.004 \\
		& RGB+Depth  & \textbf{0.639 $\pm$ 0.003} & \textbf{0.388 $\pm$ 0.005} & \textbf{0.210 $\pm$ 0.004} & \textbf{0.225 $\pm$ 0.004} \\
		\hline
	\end{tabular}
	\label{table:modelversus}
\end{table*}

\begin{table*}[!htbp]
	\centering
	\caption{Ablation study on the ORL, Still DB, and RGB-D Object datasets (mean $\pm$ standard deviation). Higher values indicate better performance.}
	\begin{tabular}{|c|c|c|c|c|c|}
		\hline
		Datasets & Methods & NMI & ACC & ARI & F-measure\\
		\hline\hline
		\multirow{3}*{ORL}
		& U-MvDSCN & 0.892 $\pm$ 0.005 & 0.770 $\pm$ 0.009 & 0.677 $\pm$ 0.004 & 0.685 $\pm$ 0.015 \\
		& D-MvDSCN & 0.885 $\pm$ 0.003 & 0.778 $\pm$ 0.008 & 0.682 $\pm$ 0.006 & 0.690 $\pm$ 0.013 \\
		& MvDSCN & \textbf{0.943 $\pm$ 0.002} & \textbf{0.870 $\pm$ 0.006} & \textbf{0.819 $\pm$ 0.001} & \textbf{0.834 $\pm$ 0.012} \\
		\hline
		\multirow{3}*{Still DB}
		& U-MvDSCN & 0.203 $\pm$ 0.006 & 0.356 $\pm$ 0.009 & 0.131 $\pm$ 0.008 & 0.284 $\pm$ 0.007 \\
		& D-MvDSCN & 0.173 $\pm$ 0.008 & 0.302 $\pm$ 0.005 & 0.134 $\pm$ 0.007 & 0.295 $\pm$ 0.008 \\
		& MvDSCN & \textbf{0.245 $\pm$ 0.020} & \textbf{0.377 $\pm$ 0.023} & \textbf{0.169 $\pm$ 0.003} & \textbf{0.320 $\pm$ 0.015} \\
		\hline
		\multirow{3}*{RGB-D Object}
		& U-MvDSCN & 0.594 $\pm$ 0.004 & 0.343 $\pm$ 0.007 & 0.190 $\pm$ 0.006 & 0.205 $\pm$ 0.006 \\
		& D-MvDSCN & 0.593 $\pm$ 0.005 & 0.350 $\pm$ 0.006 & 0.192 $\pm$ 0.006 & 0.197 $\pm$ 0.006 \\
		& MvDSCN & \textbf{0.639 $\pm$ 0.003} & \textbf{0.388 $\pm$ 0.005} & \textbf{0.210 $\pm$ 0.004} & \textbf{0.225 $\pm$ 0.004} \\
		\hline
	\end{tabular}
	\label{table:ablation}
\end{table*}
\subsection{Results of Multi-modal Subspace Clustering}
For the multi-modal experiments, we used pre-trained deep autoencoders to extract features to compare shallow methods.
There were $4,096$ features for both the RGB image and depth image.
The experimental results on the RGB-D Object dataset are presented in Table \ref{table:multi-modal}.
Compared with the shallow model, we used a deep convolutional autoencoder for hidden representation, which will obtain better feature information. Compared with the shallow models that extract deep features and then conduct subspace clustering, our proposed MvDSCN joints feature learning and subspace clustering together, and multi-view relations affect both parts.
Our method outperforms all other competitors.
Compared with the second-best method, we improved performance by approximately $1.2\%$ in NMI, $1.5\%$ in ACC, and $0.8\%$ in ARI.
{We also evaluated the clustering methods on the CUB dataset.
As shown in Table \ref{table:cub}, the released implementations of some traditional algorithms cannot perform clustering in this large-scale environment.
We used ResNet50 to extract features with 2,048 dimensions for the RGB modality and 312 binary attributes for the other modality.
On CUB, our method gains improvements of approximately $0.6\%$ and $0.2\%$ over the second-best method in terms of NMI and ARI, respectively.
Furthermore, our method gains substantial improvements of approximately $8.6\%$, $13.3\%$, $17.9\%$, and $19.6\%$ over the best non-deep method (LMSC) in terms of NMI, ACC, ARI, and F-measure, respectively.
}

In contrast to the shallow models that extract deep features and then perform subspace clustering, the MvDSCN joins feature learning and subspace clustering so that the multi-view relations affect both parts. Hence, MvDSCN achieves outstanding performance.

\textbf{Single modality versus multiple modalities}.
As shown in Table \ref{table:modelversus}, our method substantially outperforms subspace clustering with only a single modality of data, which further demonstrates the superiority of multi-modal fusion.
Overall, better results are obtained using the RGB modality than the depth modality.
Compared to subspace clustering with only the RGB modality, we improved the clustering performance by $5.0\%$ in NMI, $4.9\%$ in ACC, $4.7\%$ in ARI, and $4.6\%$ in F-measure when we fuse them using the MvDSCN.

\begin{figure*}[!htbp]
	\tiny
	\centering
	\subfigure[$\lambda_1$]{
		\begin{minipage}{0.242\linewidth}
			\centering	\includegraphics[width=1\linewidth]{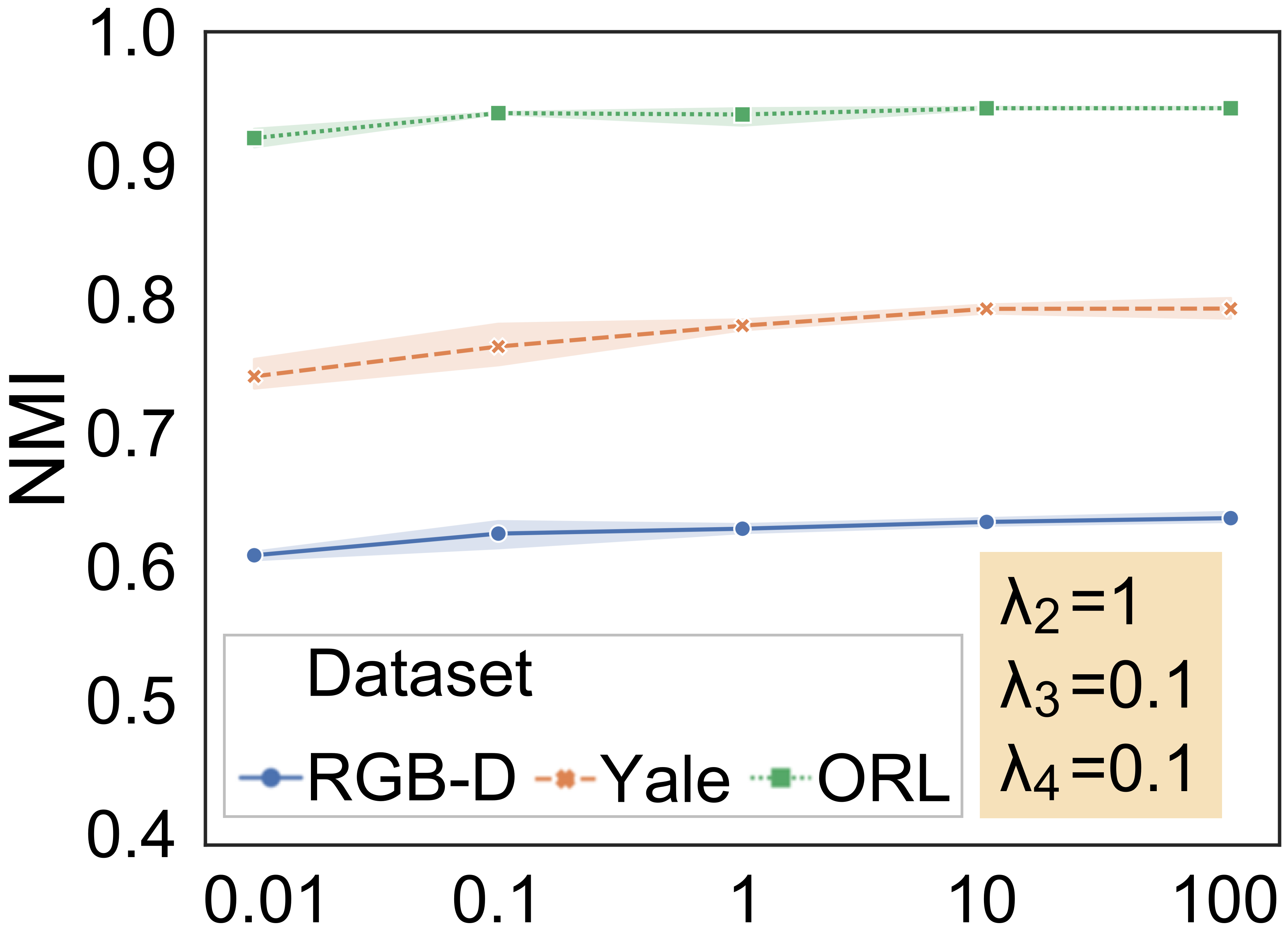}
	\end{minipage}}
	\subfigure[$\lambda_2$]{
		\begin{minipage}{0.242\linewidth}
			\centering
			\includegraphics[width=1\linewidth]{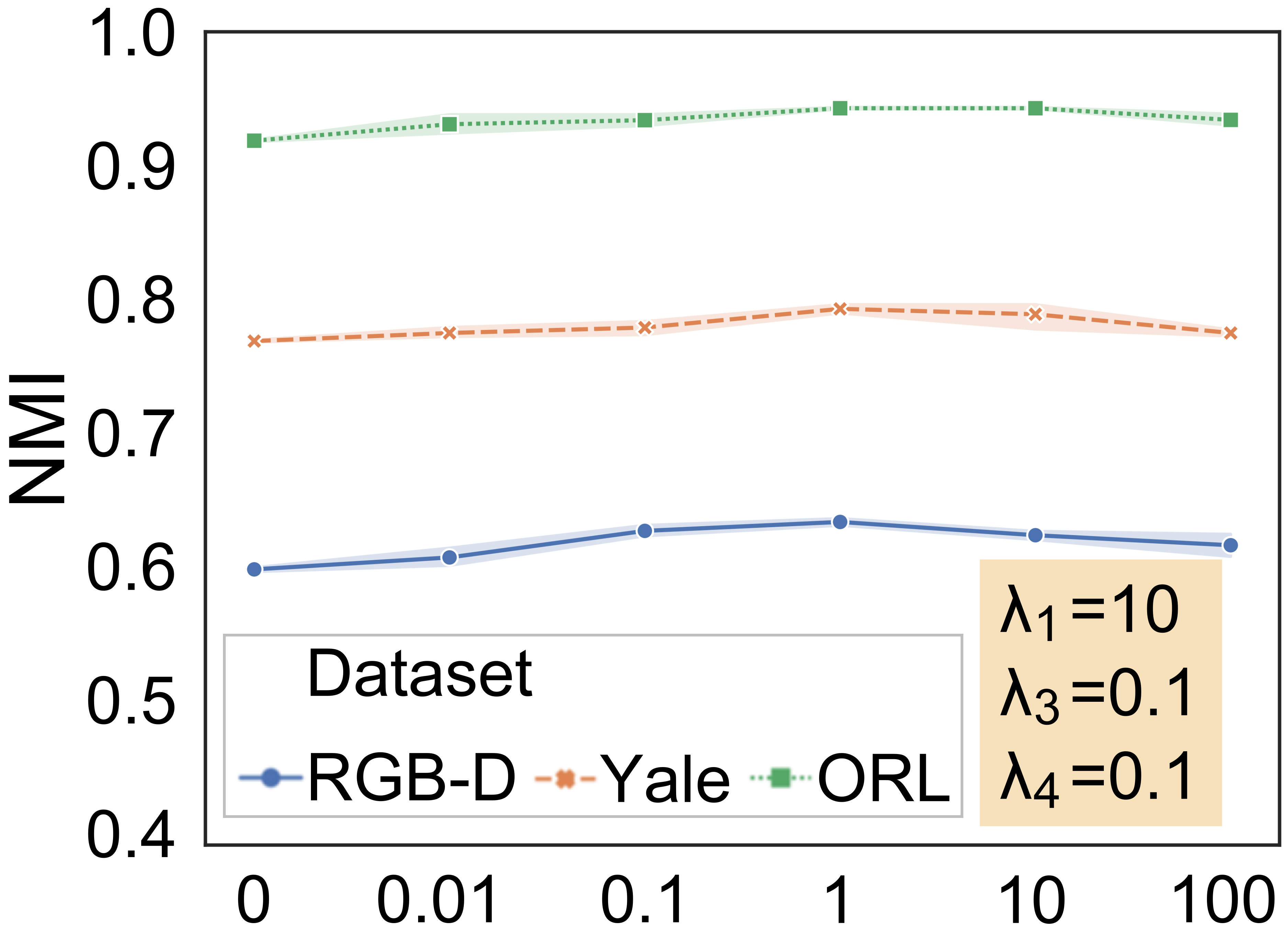}
		\end{minipage}
	}
	\subfigure[$\lambda_3$]{
		\begin{minipage}{0.242\linewidth}
			\centering
			\includegraphics[width=1\linewidth]{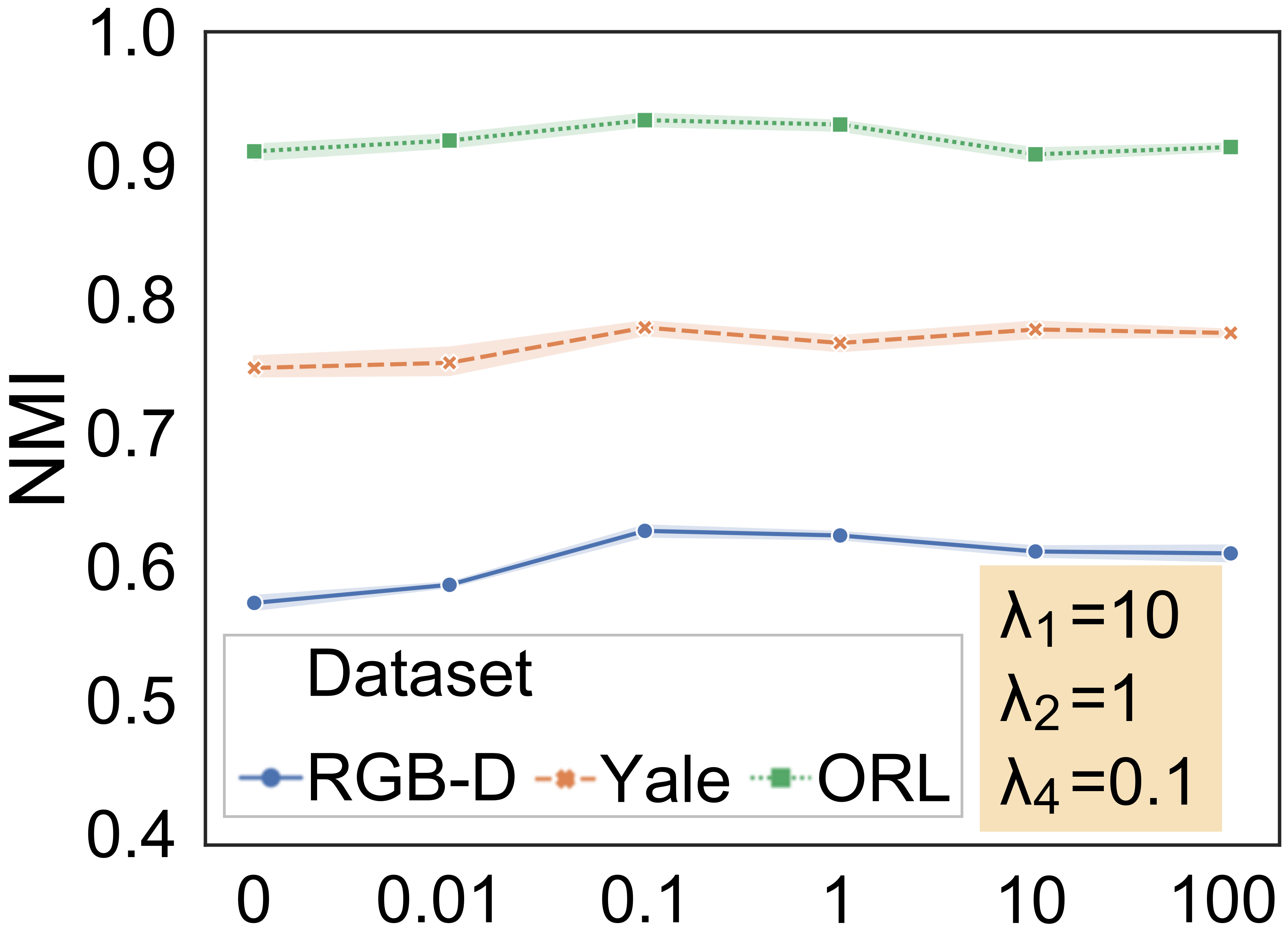}
	\end{minipage}}
	\subfigure[$\lambda_4$]{
		\begin{minipage}{0.242\linewidth}
			\centering
			\includegraphics[width=1\linewidth]{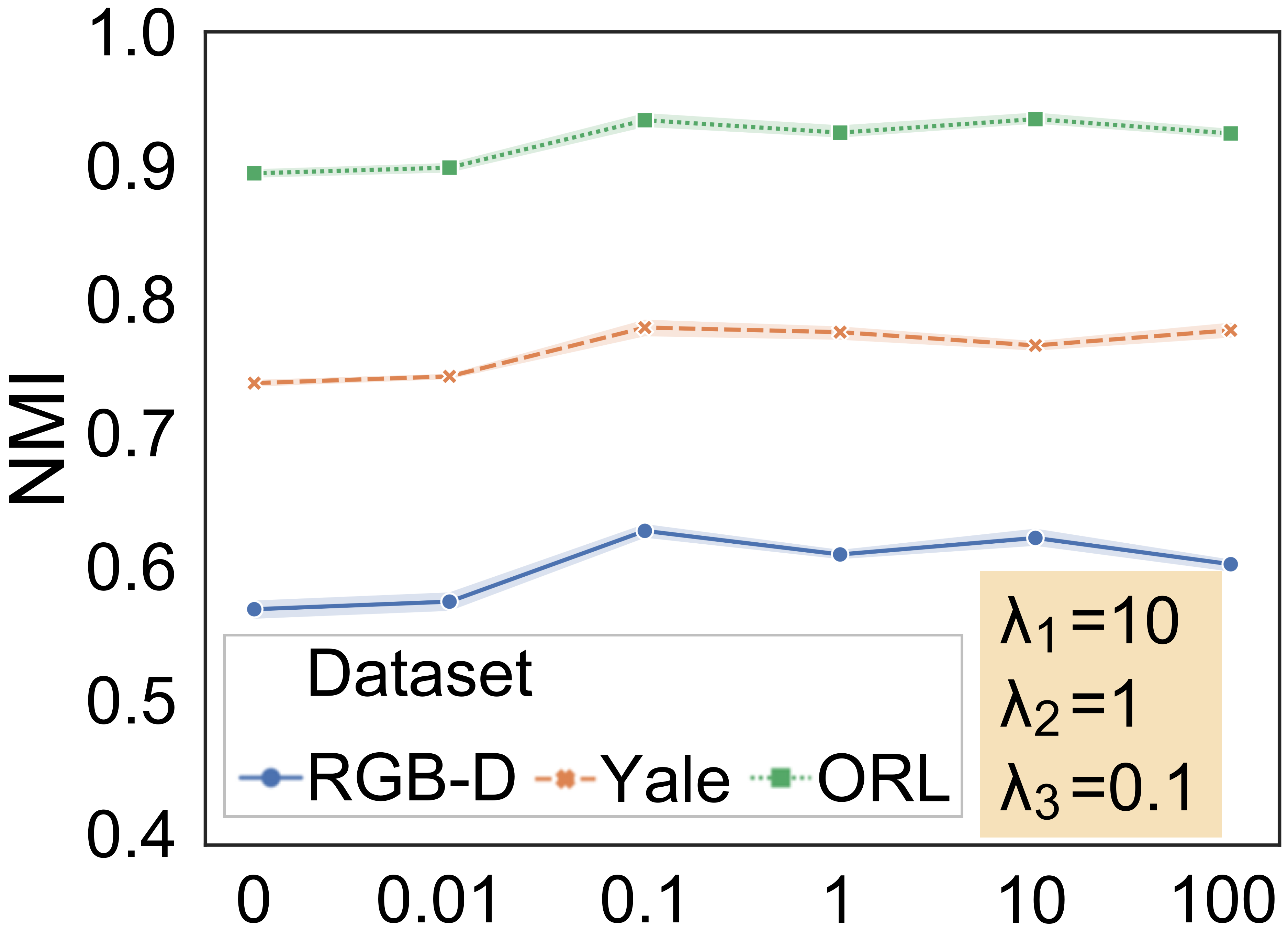}
	\end{minipage}}
	\caption{{Effect of different parameters on MvDSCN learning.}}
	\label{fig:sen}
\end{figure*}

\subsection{Multiple Features versus Multiple Modalities}
The MvDSCN is developed to adopt an end-to-end manner, in contrast to existing multi-view clustering methods with handcrafted features.
Compared with its performance on datasets with multiple features, the performance of MvDSCN is better on datasets with multiple modalities.
As shown in Table \ref{table:multiview}, the MvDSCN achieves a performance that is comparable with that of FMR.
However, the results on multi-modal datasets (the RGB-D Object and CUB datasets) show that MvDSCN substantially outperforms FMR.
That is because MvDSCN joins feature learning and affinity matrix learning, whereas FMR only learns the affinity matrix for the extracted features.
With handcrafted features as the input, MvDSCN does not perform as well as when the raw data are used as the input because handcrafted features lose some information and break the original structure of the raw data.
Hence, MvDSCN is more suitable for multi-modal applications.

\subsection{Ablation Study}
To verify the effectiveness of the diversity and universality regularizations, we conducted an ablation study on the proposed model. Results are shown in Table \ref{table:ablation}, where U-MvDSCN represents the proposed model without diversity regularization, whereas D-MvDSCN represents the one without universality regularization. As presented in Table \ref{table:ablation}, MvDSCN substantially outperforms U-MvDSCN, which quantitatively indicates that the diversity regularization that learns the multi-view complementary information benefits the clustering performance. Moreover, our proposed method performs better than D-MvDSCN, demonstrating that the universality regularization that learns common representations of different views is also significant for improving the clustering performance. According to Table \ref{table:ablation}, it can be seen that Unet and Dnet have contributed significantly to the clustering performance, which demonstrates that embedding multi-view relations into feature learning can improve the performance of clustering. In other words, diversity and universality are both crucial for multi-view learning, and the explicit decoupling design in our method is promising to enhance the learning of both aspects simultaneously.

\begin{figure}[!htbp]
	\tiny
	\centering
	\subfigure[ORL-NMI]{
		\begin{minipage}{0.49\linewidth}
			\centering	\includegraphics[width=1\linewidth]{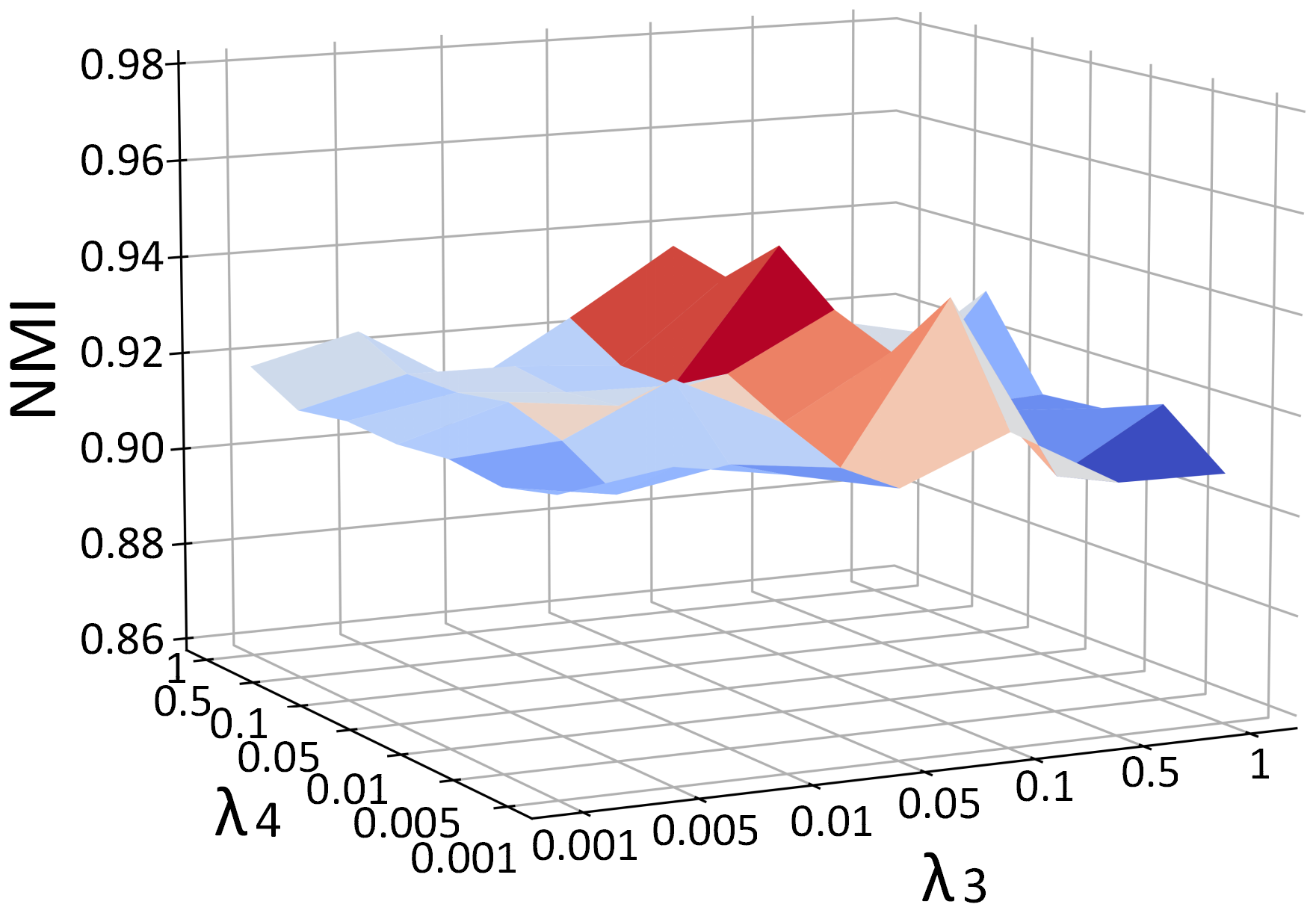}
	\end{minipage}}
	\subfigure[ORL-ACC]{
		\begin{minipage}{0.49\linewidth}
			\centering	\includegraphics[width=1\linewidth]{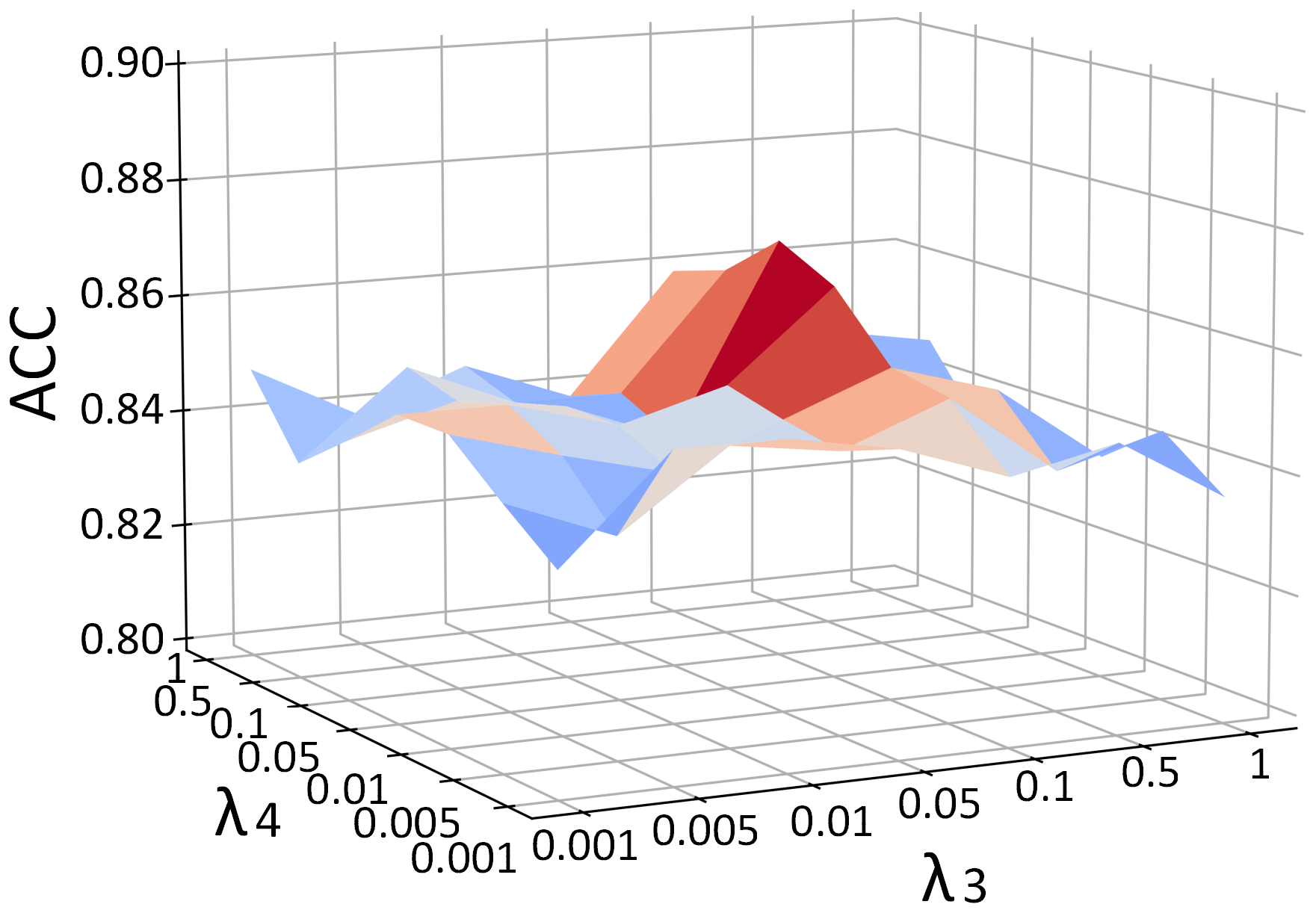}
	\end{minipage}}
\\
	\subfigure[RGB-D Object-NMI]{
		\begin{minipage}{0.49\linewidth}
			\centering  \includegraphics[width=1\linewidth]{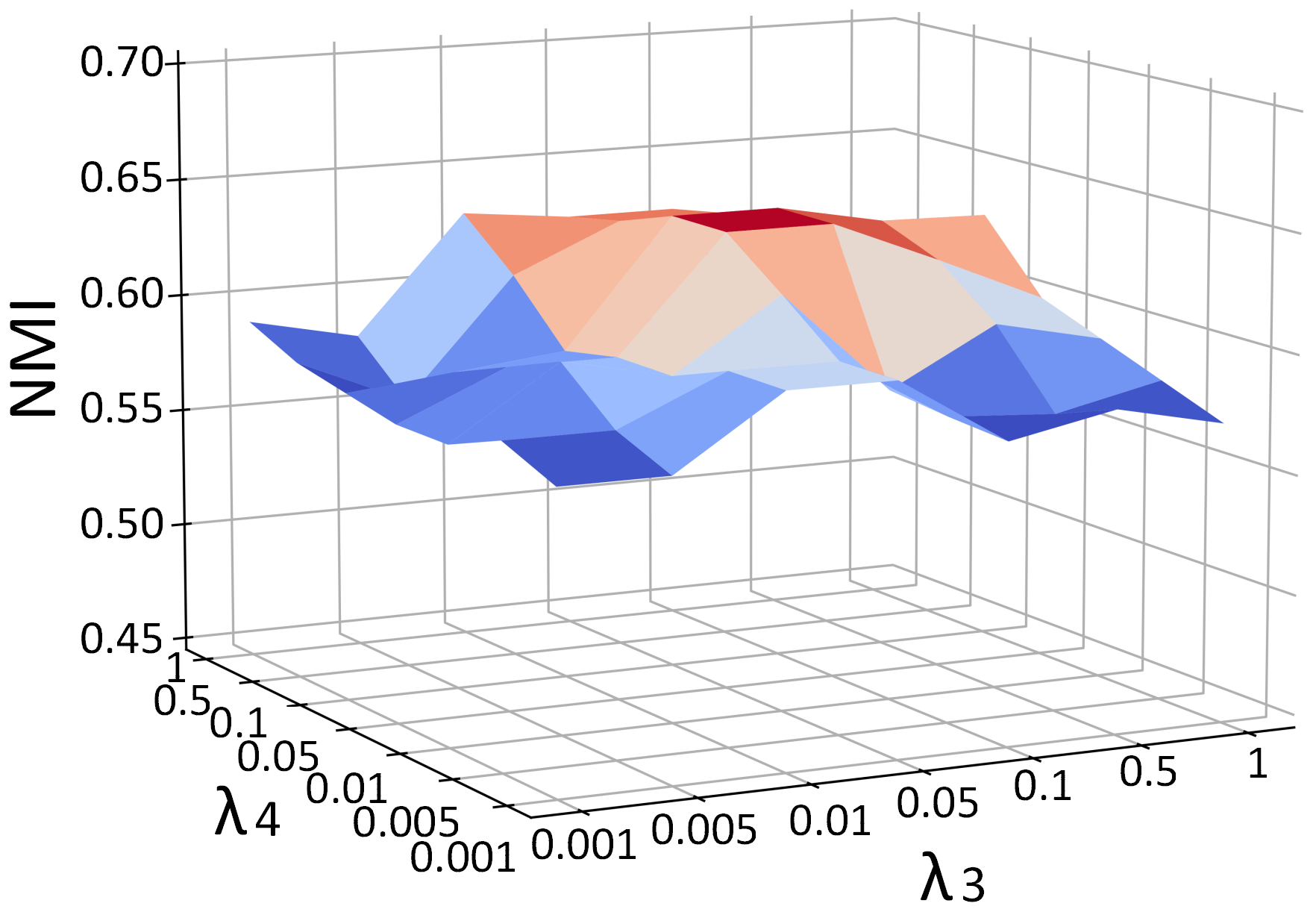}
	\end{minipage}}
	\subfigure[RGB-D Object-ACC]{
		\begin{minipage}{0.49\linewidth}
			\centering  \includegraphics[width=1\linewidth]{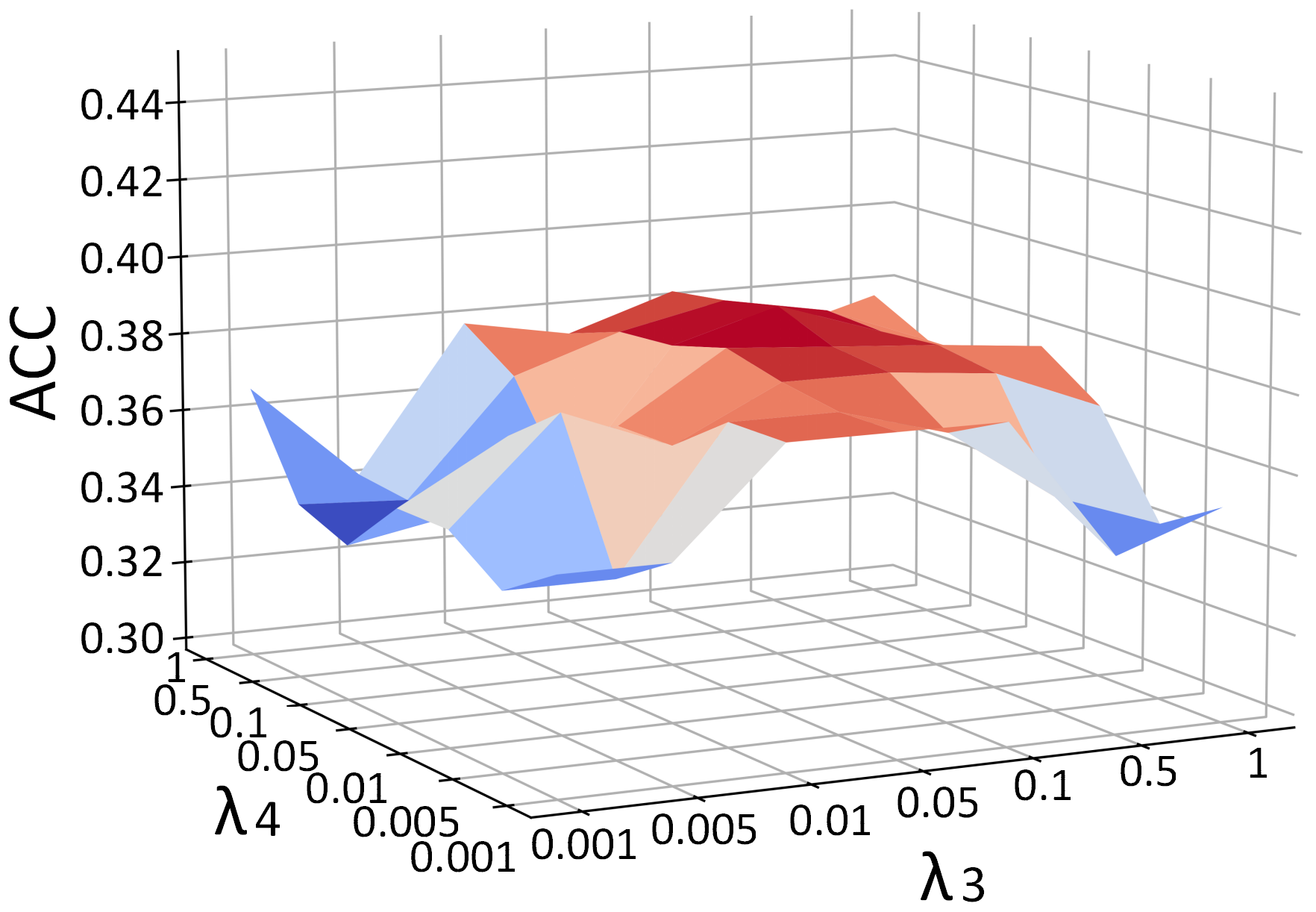}
	\end{minipage}}
	\caption{Sensitivity analysis.}
	\label{fig:sensitivity_test}
\end{figure}

\subsection{Parameter Analysis}
To analyze the impact of the parameters on the clustering performance of the MvDSCN, we plotted the performance of MvDSCN in Fig. \ref{fig:sen}
with four different hyper-parameters, $\lambda_1$, $\lambda_2$, $\lambda_3$, and $\lambda_4$.
Parameter $\lambda_1$ is used to adjust the impact of the self-representation loss.
Parameter $\lambda_2$ adjusts the impact of the $l_p$-norm regularization.
Parameter $\lambda_3$ controls the degree of the universality regularizer, whereas $\lambda_4$ controls the degree of the diversity regularizer.
We fixed three parameters and analyzed the impact of the remaining parameter.
In addition to $\lambda_1$, we explored the effect of parameter changes from $0$ to $100$ on final performance.
The results show that the clustering performance grows with the value of $\lambda_1$ and is stable if $\lambda_1$ is above $10$.
The best performance is achieved across different datasets when $\lambda_2$ is set to $1$.
Similarly, we can observe that when $\lambda_3$ and $\lambda_4$ are set as $0.1$, the MvDSCN achieves superior performance. As shown in Fig. \ref{fig:sen}, all four hyper-parameters contribute to improving the performance of the model, which shows that such a decoupled-based design allows different regular terms to learn knowledge in a more targeted manner and avoids interfering with each other.

In Table \ref{table:ablation} and Fig. \ref{fig:sen}, it can be seen that diversity regularization and universality regularization both contribute to improving the performance of the model. 
In Fig. \ref{fig:sen}, we performed extensive experiments to discuss the influence of hyper-parameters in the objective function. 
Results are shown in Fig. \ref{fig:sen}, and we can make the following observations. 
The clustering performance is stable with different values of $\lambda_1$ and $\lambda_2$ while relatively more sensitive with different values of $\lambda_3$ and $\lambda_4$, which control the core modules of our model, Unet and Dnet. We further conducted a sensitivity analysis for the regularization parameter of diversity ($\lambda_3$) and universality ($\lambda_4$) varying from $0.001$ to $1$. As shown in Fig. \ref{fig:sensitivity_test}, our method performs more stable when the values of $\lambda_3$ and $\lambda_4$ are 0.1.

In sum, all four hyper-parameters contribute to the enhancement of the proposed model in terms of multi-view clustering performance. For all datasets, we set the values of four hyper-parameters to 10, 1, 0.1, and 0.1, respectively, because this combination of parameters generally resulted in the best performance. 

\subsection{Convergence Analysis}

\begin{figure}[!htbp]
	\centering
 	\includegraphics[width=0.8\linewidth]{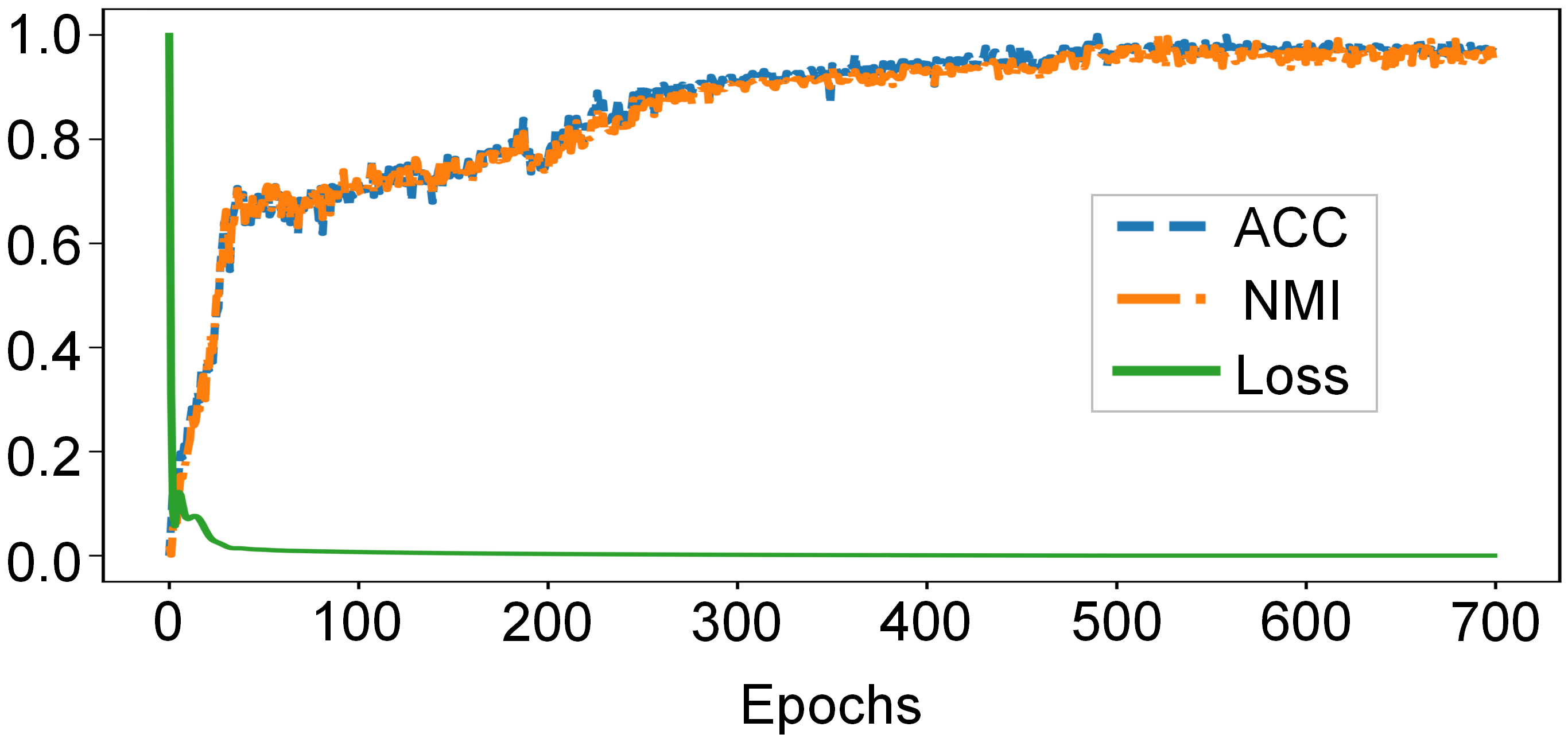}
	\caption{{ Loss and clustering performance (NMI and ACC) of the MvDSCN with respect to training epochs.}}
	\label{fig:converage}
\end{figure}

To empirically analyze the convergence of the MvDSCN, in Fig. \ref{fig:converage}, we show the relationship between the loss of the MvDSCN and the clustering performance on the ORL dataset. The reported values in this figure are normalized to a range between zero and one. As can be seen from the figure, the loss decreases rapidly in a few epochs.
The clustering performance increases substantially in the first few epochs and then grows slowly. Similar results can be observed on other datasets.

\section{Conclusions}
\label{s5}
{
In this paper, MvDSCN unifies multiple backbones to boost clustering performance and avoid the need for model selection.
The proposed method learns multi-view self-representation in an end-to-end manner by combining a convolutional autoencoder with self-representation.
It consists of two sub-networks and embeds multi-view relations into feature learning to obtain a better clustering effect, which is connected by diversity and universality regularizers.
Extensive experiments show that our method achieves outstanding performance.
In the future, we can extend the study to multi-modal continual learning and dynamic multi-modality.}

\bibliographystyle{IEEEtran}
\bibliography{egbib}

\begin{thebibliography}{10}
\providecommand{\url}[1]{#1}
\csname url@samestyle\endcsname
\providecommand{\newblock}{\relax}
\providecommand{\bibinfo}[2]{#2}
\providecommand{\BIBentrySTDinterwordspacing}{\spaceskip=0pt\relax}
\providecommand{\BIBentryALTinterwordstretchfactor}{4}
\providecommand{\BIBentryALTinterwordspacing}{\spaceskip=\fontdimen2\font plus
\BIBentryALTinterwordstretchfactor\fontdimen3\font minus
  \fontdimen4\font\relax}
\providecommand{\BIBforeignlanguage}[2]{{%
\expandafter\ifx\csname l@#1\endcsname\relax
\typeout{** WARNING: IEEEtran.bst: No hyphenation pattern has been}%
\typeout{** loaded for the language `#1'. Using the pattern for}%
\typeout{** the default language instead.}%
\else
\language=\csname l@#1\endcsname
\fi
#2}}
\providecommand{\BIBdecl}{\relax}
\BIBdecl

\bibitem{Guangcan2013Robust}
G.~Liu, Z.~Lin, S.~Yan, J.~Sun, Y.~Yu, and Y.~Ma, ``Robust recovery of subspace
  structures by low-rank representation,'' \emph{IEEE Transactions on Pattern
  Analysis and Machine Intelligence}, vol.~35, no.~1, pp. 171--184, 2012.

\bibitem{ji2017deep}
P.~Ji, T.~Zhang, H.~Li, M.~Salzmann, and I.~Reid, ``Deep subspace clustering
  networks,'' \emph{NeurIPS}, vol.~30, pp. 1--10, 2017.

\bibitem{rw_subspace_tcyb_2}
Y.~Qin, G.~Feng, Y.~Ren, and X.~Zhang, ``Consistency-induced multiview subspace
  clustering,'' \emph{IEEE Transactions on Cybernetics}, vol.~53, no.~2, pp.
  832--844, 2023.

\bibitem{Lu2018Subspace}
C.~Lu, J.~Feng, Z.~Lin, T.~Mei, and S.~Yan, ``Subspace clustering by block
  diagonal representation,'' \emph{IEEE Transactions on Pattern Analysis and
  Machine Intelligence}, vol.~41, no.~2, pp. 487--501, 2018.

\bibitem{rw_wang2021hierarchical_tcyb_6}
Y.~Wang, Z.~Wang, Q.~Hu, Y.~Zhou, and H.~Su, ``Hierarchical semantic risk
  minimization for large-scale classification,'' \emph{IEEE Transactions on
  Cybernetics}, vol.~52, no.~9, pp. 9546--9558, 2022.

\bibitem{Jiang2018WhenTL}
Y.~Jiang, Z.~Yang, Q.~Xu, X.~Cao, and Q.~Huang, ``When to learn what: Deep
  cognitive subspace clustering,'' in \emph{ACM MM}, 2018, pp. 718--726.

\bibitem{luo2017simple}
M.~Luo, X.~Chang, Z.~Li, L.~Nie, A.~G. Hauptmann, and Q.~Zheng, ``Simple to
  complex cross-modal learning to rank,'' \emph{Computer Vision and Image
  Understanding}, vol. 163, pp. 67--77, 2017.

\bibitem{kim2015dasc}
S.~Kim, D.~Min, B.~Ham, S.~Ryu, M.~N. Do, and K.~Sohn, ``Dasc: Dense adaptive
  self-correlation descriptor for multi-modal and multi-spectral
  correspondence,'' in \emph{CVPR}, 2015, pp. 2103--2112.

\bibitem{chen2017multi}
X.~Chen, H.~Ma, J.~Wan, B.~Li, and T.~Xia, ``Multi-view 3d object detection
  network for autonomous driving,'' in \emph{CVPR}, 2017, pp. 1907--1915.

\bibitem{Zhao2017MultiViewCV}
H.~Zhao, Z.~Ding, and Y.~Fu, ``Multi-view clustering via deep matrix
  factorization,'' in \emph{AAAI}, 2017, pp. 2921–--2927.

\bibitem{li2021self}
K.~Li, H.~Liu, Y.~Zhang, K.~Li, and Y.~Fu, ``Self-guided deep multiview
  subspace clustering via consensus affinity regularization,'' \emph{IEEE
  Transactions on Cybernetics}, vol.~52, no.~12, pp. 12\,734--12\,744, 2022.

\bibitem{zhou2019dual}
T.~Zhou, C.~Zhang, X.~Peng, H.~Bhaskar, and J.~Yang, ``Dual shared-specific
  multiview subspace clustering,'' \emph{IEEE Transactions on Cybernetics},
  vol.~50, no.~8, pp. 3517--3530, 2020.

\bibitem{tang2021one}
Y.~Tang, Y.~Xie, C.~Zhang, Z.~Zhang, and W.~Zhang, ``One-step multiview
  subspace segmentation via joint skinny tensor learning and latent
  clustering,'' \emph{IEEE Transactions on Cybernetics}, vol.~52, no.~9, pp.
  9179--9193, 2022.

\bibitem{xing2019correntropy}
L.~Xing, B.~Chen, S.~Du, Y.~Gu, and N.~Zheng, ``Correntropy-based multiview
  subspace clustering,'' \emph{IEEE Transactions on Cybernetics}, vol.~51,
  no.~6, pp. 3298--3311, 2021.

\bibitem{chen2020multiview}
M.-S. Chen, L.~Huang, C.-D. Wang, D.~Huang, and S.~Y. Philip, ``Multiview
  subspace clustering with grouping effect,'' \emph{IEEE Transactions on
  Cybernetics}, vol.~52, no.~8, pp. 7655--7668, 2022.

\bibitem{Zhang2017LatentMS}
C.~Zhang, Q.~Hu, H.~Fu, P.~Zhu, and X.~Cao, ``Latent multi-view subspace
  clustering,'' in \emph{CVPR}, 2017, pp. 4279--4287.

\bibitem{rw_subspace_tcyb_3}
Z.~Yu, G.~Zhang, J.~Chen, H.~Chen, D.~Zhang, Q.~Yang, and J.~Shao, ``Toward
  noise-resistant graph embedding with subspace clustering information,''
  \emph{IEEE Transactions on Cybernetics}, vol.~53, no.~5, pp. 2980--2992,
  2023.

\bibitem{rw_low_rank_tcyb_1}
Z.~Fu, Y.~Zhao, D.~Chang, Y.~Wang, and J.~Wen, ``Latent low-rank representation
  with weighted distance penalty for clustering,'' \emph{IEEE Transactions on
  Cybernetics}, vol.~53, no.~11, pp. 6870--6882, 2023.

\bibitem{rw_low_rank_conference_1}
C.~Zhang, H.~Li, W.~Lv, Z.~Huang, Y.~Gao, and C.~Chen, ``Enhanced tensor
  low-rank and sparse representation recovery for incomplete multi-view
  clustering,'' in \emph{AAAI}, vol.~37, no.~9, 2023, pp. 11\,174--11\,182.

\bibitem{yu2020clustering}
Z.~Yu, D.~Wang, X.-B. Meng, and C.~P. Chen, ``Clustering ensemble based on
  hybrid multiview clustering,'' \emph{IEEE Transactions on Cybernetics},
  vol.~52, no.~7, pp. 6518--6530, 2022.

\bibitem{rw_mv_survey_1}
U.~Fang, M.~Li, J.~Li, L.~Gao, T.~Jia, and Y.~Zhang, ``A comprehensive survey
  on multi-view clustering,'' \emph{IEEE Transactions on Knowledge and Data
  Engineering}, vol.~35, no.~12, pp. 12\,350--12\,368, 2023.

\bibitem{Li2017MultiviewGL}
S.~Li, H.~Liu, Z.~Tao, and Y.~Fu, ``Multi-view graph learning with adaptive
  label propagation,'' in \emph{IEEE BigData}, 2017, pp. 110--115.

\bibitem{chen2021multiview}
J.~Chen, S.~Yang, H.~Mao, and C.~Fahy, ``Multiview subspace clustering using
  low-rank representation,'' \emph{IEEE Transactions on Cybernetics}, vol.~52,
  no.~11, pp. 12\,364--12\,378, 2022.

\bibitem{rw_spectral_tcyb_5}
D.~Shi, L.~Zhu, J.~Li, Z.~Cheng, and Z.~Zhang, ``Flexible multiview spectral
  clustering with self-adaptation,'' \emph{IEEE Transactions on Cybernetics},
  vol.~53, no.~4, pp. 2586--2599, 2023.

\bibitem{kang2021structured}
Z.~Kang, Z.~Lin, X.~Zhu, and W.~Xu, ``Structured graph learning for scalable
  subspace clustering: From single view to multiview,'' \emph{IEEE Transactions
  on Cybernetics}, vol.~52, no.~9, pp. 8976--8986, 2022.

\bibitem{Tao2017FromEC}
Z.~Tao, H.~Liu, S.~Li, Z.~Ding, and Y.~Fu, ``From ensemble clustering to
  multi-view clustering,'' in \emph{IJCAI}, 2017, pp. 2843--2849.

\bibitem{Tao2019MarginalizedME}
{Tao, Zhiqiang and Liu, Hongfu and Li, Sheng and Ding, Zhengming and Fu, Yun},
  ``Marginalized multiview ensemble clustering,'' \emph{IEEE Transactions on
  Neural Networks and Learning Systems}, vol.~31, no.~2, pp. 600--611, 2019.

\bibitem{DCP}
Y.~Lin, Y.~Gou, X.~Liu, J.~Bai, J.~Lv, and X.~Peng, ``Dual contrastive
  prediction for incomplete multi-view representation learning,'' \emph{IEEE
  Transactions on Pattern Analysis and Machine Intelligence}, vol.~45, no.~4,
  pp. 4447--4461, 2023.

\bibitem{SURE}
M.~Yang, Y.~Li, P.~Hu, J.~Bai, J.~Lv, and X.~Peng, ``Robust multi-view
  clustering with incomplete information,'' \emph{IEEE Transactions on Pattern
  Analysis and Machine Intelligence}, vol.~45, no.~1, pp. 1055--1069, 2023.

\bibitem{rw_tensor_tcyb_4}
J.-Q. Lin, M.-S. Chen, C.-D. Wang, and H.~Zhang, ``A tensor approach for
  uncoupled multiview clustering,'' \emph{IEEE Transactions on Cybernetics},
  vol.~54, no.~2, pp. 1236--1249, 2024.

\bibitem{rw_aggregation_conference_2}
W.~Yan, Y.~Zhang, C.~Lv, C.~Tang, G.~Yue, L.~Liao, and W.~Lin, ``Gcfagg: Global
  and cross-view feature aggregation for multi-view clustering,'' in
  \emph{CVPR}, 2023, pp. 19\,863--19\,872.

\bibitem{ZhangGeneralized}
C.~Zhang, H.~Fu, Q.~Hu, X.~Cao, Y.~Xie, D.~Tao, and D.~Xu, ``Generalized latent
  multi-view subspace clustering,'' \emph{IEEE Transactions on Pattern Analysis
  and Machine Intelligence}, vol.~42, no.~1, pp. 86--99, 2018.

\bibitem{buades2005non}
A.~Buades, B.~Coll, and J.-M. Morel, ``A non-local algorithm for image
  denoising,'' in \emph{CVPR}, vol.~2, 2005, pp. 60--65.

\bibitem{zhu2017subspace}
P.~Zhu, W.~Zhu, Q.~Hu, C.~Zhang, and W.~Zuo, ``Subspace clustering guided
  unsupervised feature selection,'' \emph{Pattern Recognition}, vol.~66, pp.
  364--374, 2017.

\bibitem{wang2018non}
X.~Wang, R.~Girshick, A.~Gupta, and K.~He, ``Non-local neural networks,'' in
  \emph{CVPR}, 2018, pp. 7794--7803.

\bibitem{Hinton2006ReducingTD}
G.~E. Hinton and R.~R. Salakhutdinov, ``Reducing the dimensionality of data
  with neural networks,'' \emph{Science}, vol. 313, no. 5786, pp. 504--507,
  2006.

\bibitem{Vincent2010StackedDA}
P.~Vincent, H.~Larochelle, I.~Lajoie, Y.~Bengio, P.-A. Manzagol, and L.~Bottou,
  ``Stacked denoising autoencoders: Learning useful representations in a deep
  network with a local denoising criterion.'' \emph{Journal of Machine Learning
  Research}, vol.~11, no.~12, pp. 3371--3408, 2010.

\bibitem{Xie2016UnsupervisedDE}
J.~Xie, R.~Girshick, and A.~Farhadi, ``Unsupervised deep embedding for
  clustering analysis,'' in \emph{ICML}, 2016, pp. 478--487.

\bibitem{Peng2016DeepSC}
X.~Peng, S.~Xiao, J.~Feng, W.-Y. Yau, and Z.~Yi, ``Deep subspace clustering
  with sparsity prior.'' in \emph{IJCAI}, 2016, pp. 1925--1931.

\bibitem{Masci2011StackedCA}
J.~Masci, U.~Meier, D.~Cire{\c{s}}an, and J.~Schmidhuber, ``Stacked
  convolutional auto-encoders for hierarchical feature extraction,'' in
  \emph{ICANN}, 2011, pp. 52--59.

\bibitem{Nguyen2017PlugP}
A.~Nguyen, J.~Clune, Y.~Bengio, A.~Dosovitskiy, and J.~Yosinski, ``Plug \& play
  generative networks: Conditional iterative generation of images in latent
  space,'' in \emph{CVPR}, 2017, pp. 4467--4477.

\bibitem{Nair2010RectifiedLU}
V.~Nair and G.~E. Hinton, ``Rectified linear units improve restricted boltzmann
  machines,'' in \emph{ICML}, 2010, pp. 807--817.

\bibitem{gretton2005measuring}
A.~Gretton, O.~Bousquet, A.~Smola, and B.~Sch{\"o}lkopf, ``Measuring
  statistical dependence with hilbert-schmidt norms,'' in \emph{ALT}, 2005, pp.
  63--77.

\bibitem{gatys2015neural}
L.~A. Gatys, A.~S. Ecker, and M.~Bethge, ``A neural algorithm of artistic
  style,'' \emph{arXiv preprint arXiv:1508.06576}, 2015.

\bibitem{Cao2015DiversityinducedMS}
X.~Cao, C.~Zhang, H.~Fu, S.~Liu, and H.~Zhang, ``Diversity-induced multi-view
  subspace clustering,'' in \emph{CVPR}, 2015, pp. 586--594.

\bibitem{lu2019subspace}
C.~Lu, J.~Feng, Z.~Lin, T.~Mei, and S.~Yan, ``Subspace clustering by block
  diagonal representation,'' \emph{IEEE Transactions on Pattern Analysis and
  Machine Intelligence}, vol.~41, no.~2, pp. 487--501, 2018.

\bibitem{Kingma2015AdamAM}
D.~P. Kingma and J.~Ba, ``Adam: A method for stochastic optimization,''
  \emph{arXiv preprint arXiv:1412.6980}, 2014.

\bibitem{liaw2002classification}
A.~Liaw, M.~Wiener \emph{et~al.}, ``Classification and regression by
  randomforest,'' \emph{R News}, vol.~2, no.~3, pp. 18--22, 2002.

\bibitem{rodriguez2006rotation}
J.~J. Rodriguez, L.~I. Kuncheva, and C.~J. Alonso, ``Rotation forest: A new
  classifier ensemble method,'' \emph{IEEE Transactions on Pattern Analysis and
  Machine Intelligence}, vol.~28, no.~10, pp. 1619--1630, 2006.

\bibitem{peng2013scalable}
X.~Peng, L.~Zhang, and Z.~Yi, ``Scalable sparse subspace clustering,'' in
  \emph{CVPR}, 2013, pp. 430--437.

\bibitem{still_db}
N.~Ikizler, R.~G. Cinbis, S.~Pehlivan, and P.~Duygulu, ``Recognizing actions
  from still images,'' in \emph{ICPR}, 2008, pp. 1--4.

\bibitem{imagenet_bib}
J.~Deng, W.~Dong, R.~Socher, L.-J. Li, K.~Li, and L.~Fei-Fei, ``Imagenet: A
  large-scale hierarchical image database,'' in \emph{CVPR}, 2009, pp.
  248--255.

\bibitem{Lai2011ALH}
K.~Lai, L.~Bo, X.~Ren, and D.~Fox, ``A large-scale hierarchical multi-view
  rgb-d object dataset,'' in \emph{ICRA}, 2011, pp. 1817--1824.

\bibitem{WahCUB_200_2011}
C.~Wah, S.~Branson, P.~Welinder, P.~Perona, and S.~Belongie, ``The caltech-ucsd
  birds-200-2011 dataset,'' pp. 1--8, 2011.

\bibitem{Ng2001OnSC}
A.~Ng, M.~Jordan, and Y.~Weiss, ``On spectral clustering: Analysis and an
  algorithm,'' \emph{NeurIPS}, vol.~14, pp. 1--8, 2001.

\bibitem{Xia2014RobustMS}
R.~Xia, Y.~Pan, L.~Du, and J.~Yin, ``Robust multi-view spectral clustering via
  low-rank and sparse decomposition,'' in \emph{AAAI}, vol.~28, no.~1, 2014,
  pp. 1--7.

\bibitem{Caron2018DeepCF}
M.~Caron, P.~Bojanowski, A.~Joulin, and M.~Douze, ``Deep clustering for
  unsupervised learning of visual features,'' in \emph{ECCV}, 2018, pp.
  132--149.

\bibitem{Sa2005SpectralCW}
V.~R. De~Sa, ``Spectral clustering with two views,'' in \emph{ICML workshop on
  Learning with Multiple Views}, 2005, pp. 20--27.

\bibitem{Kumar2011CoregularizedMS}
A.~Kumar, P.~Rai, and H.~Daume, ``Co-regularized multi-view spectral
  clustering,'' \emph{NeurIPS}, vol.~24, pp. 1--9, 2011.

\bibitem{Li2019FlexibleMR}
R.~Li, C.~Zhang, Q.~Hu, P.~Zhu, and Z.~Wang, ``Flexible multi-view
  representation learning for subspace clustering.'' in \emph{IJCAI}, 2019, pp.
  2916--2922.

\bibitem{Abavisani2018DeepMS}
M.~Abavisani and V.~M. Patel, ``Deep multimodal subspace clustering networks,''
  \emph{IEEE Journal of Selected Topics in Signal Processing}, vol.~12, no.~6,
  pp. 1601--1614, 2018.

\bibitem{wang2020deep}
Q.~Wang, J.~Cheng, Q.~Gao, G.~Zhao, and L.~Jiao, ``Deep multi-view subspace
  clustering with unified and discriminative learning,'' \emph{IEEE
  Transactions on Multimedia}, vol.~23, pp. 3483--3493, 2020.

\bibitem{DPML_CPM}
C.~Zhang, Y.~Cui, Z.~Han, J.~T. Zhou, H.~Fu, and Q.~Hu, ``Deep partial
  multi-view learning,'' \emph{IEEE Transactions on Pattern Analysis and
  Machine Intelligence}, vol.~44, pp. 2402--2415, 2022.

\bibitem{wen2021unified}
J.~Wen, Z.~Zhang, Z.~Zhang, L.~Zhu, L.~Fei, B.~Zhang, and Y.~Xu, ``Unified
  tensor framework for incomplete multi-view clustering and missing-view
  inferring,'' in \emph{AAAI}, vol.~35, no.~11, 2021, pp. 10\,273--10\,281.

\end{thebibliography}
\vspace{-5mm}
\begin{IEEEbiography}[{\includegraphics[width=1in,height=1.25in,clip,keepaspectratio]{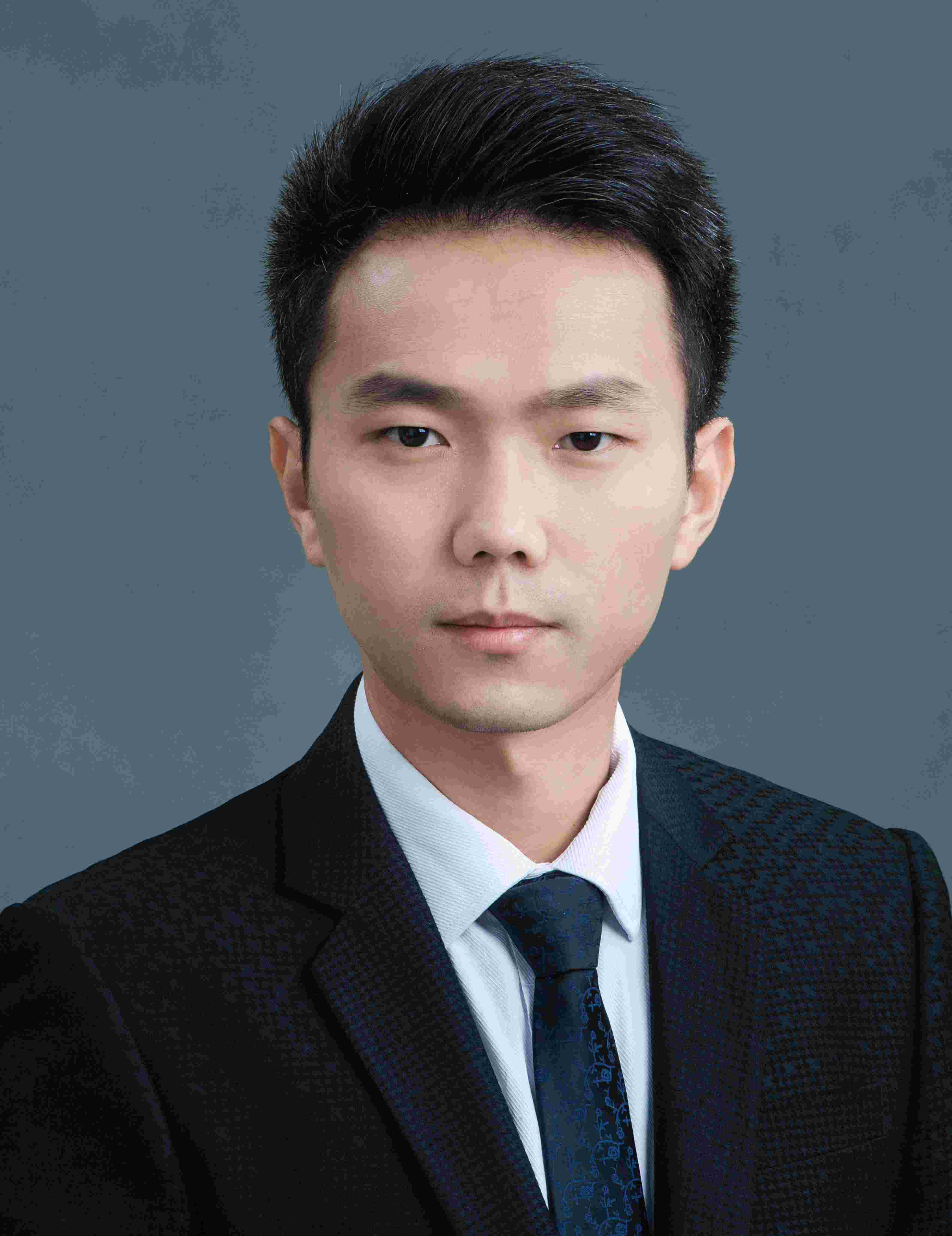}}]{Pengfei Zhu} received the Ph.D. degree from The Hong Kong Polytechnic University, Hong Kong, China, in 2015. Now he is a professor with the College of Intelligence and Computing, Tianjin University. His research interests are focused on machine learning and computer vision.
\end{IEEEbiography}
\vspace{-5mm}
\begin{IEEEbiography}[{\includegraphics[width=1in,height=1.25in,clip,keepaspectratio]{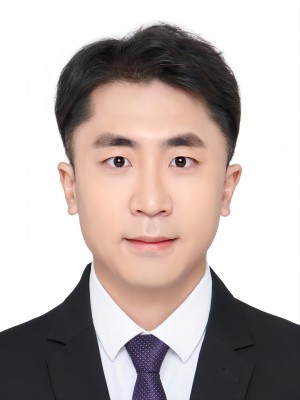}}]{Xinjie Yao} received the B.S. degree in information security from North China Electric Power University, Baoding, China, in 2018, and the M.S. degree in computer technology from China University of Petroleum (East China), Qingdao, China, in 2021. He is currently pursuing the Ph.D. degree in computer science and technology with Tianjin University, Tianjin, China. His research interests are focused on machine learning and computer vision.
\end{IEEEbiography}
\vspace{-5mm}
\begin{IEEEbiography}[{\includegraphics[width=1in,height=1.25in,clip,keepaspectratio]{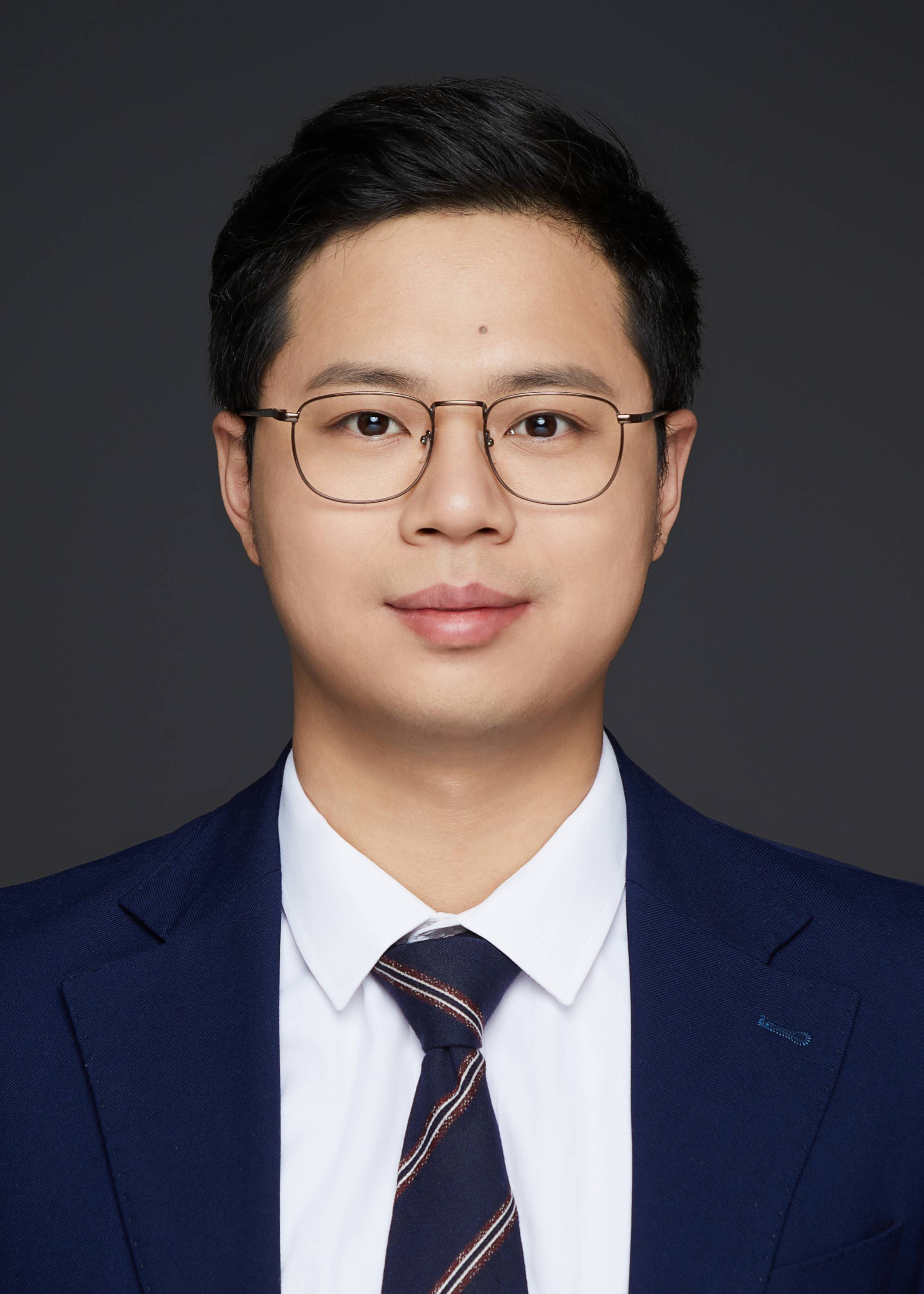}}]{Yu Wang} received the B.S. degree in communication engineering, the M.S. degree in software engineering, and the Ph.D. degree in computer applications and techniques from Tianjin University, respectively. He is currently an associate professor at Tianjin University. He has published over 40 papers including IEEE TPAMI, TKDE, etc., and has been focusing on data mining and machine learning, especially multi-granularity learning in the open and dynamic environment for computer vision and industrial applications.
\end{IEEEbiography}
\vspace{-5mm}
\begin{IEEEbiography}[{\includegraphics[width=1in,height=1.25in,clip,keepaspectratio]{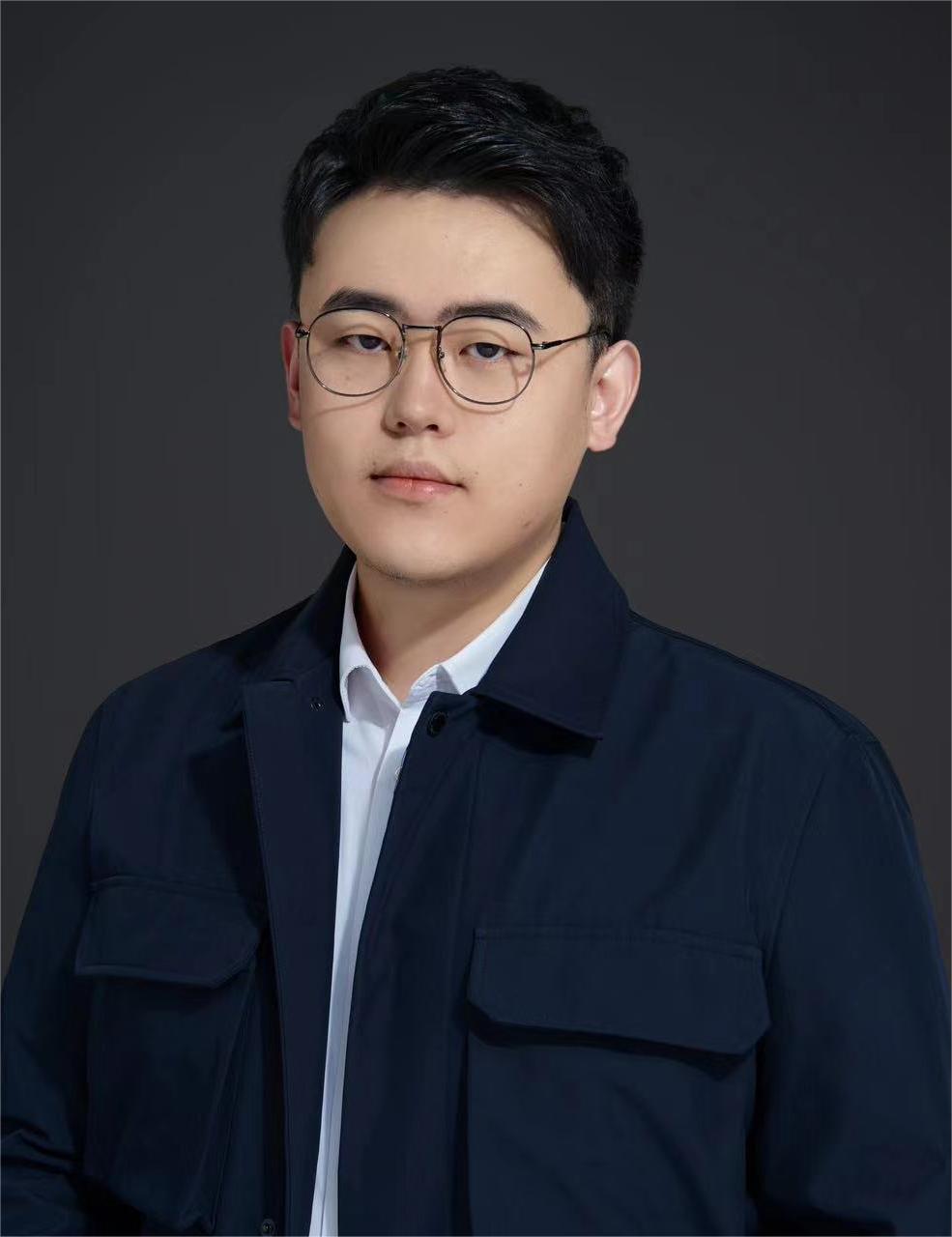}}]{Binyuan Hui} received the M.S. degree in computer science and technology from the College of Intelligence and Computing, Tianjin University, China, in 2020. His research interests include data mining, pattern recognition, computer vision, and natural language processing.
\end{IEEEbiography}
\vspace{-5mm}
\begin{IEEEbiography}[{\includegraphics[width=1in,height=1.25in,clip,keepaspectratio]{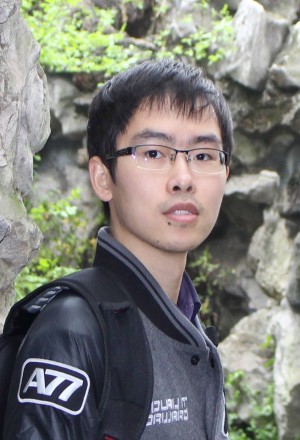}}]{Dawei Du} received the B.S. and M.S. degrees from the University of Electronic Science and Technology of China, Chengdu, China, in 2010 and 2013, respectively, and the Ph.D. degree from the University of Chinese Academy of Sciences, Beijing, China, in 2018. He is currently a postdoctoral researcher with the University at Albany, State University of New York, Albany, New York. His current research interests include visual tracking, object detection, and semantic segmentation.
\end{IEEEbiography}
\vspace{-5mm}
\begin{IEEEbiography}[{\includegraphics[width=1in,height=1.25in,clip,keepaspectratio]{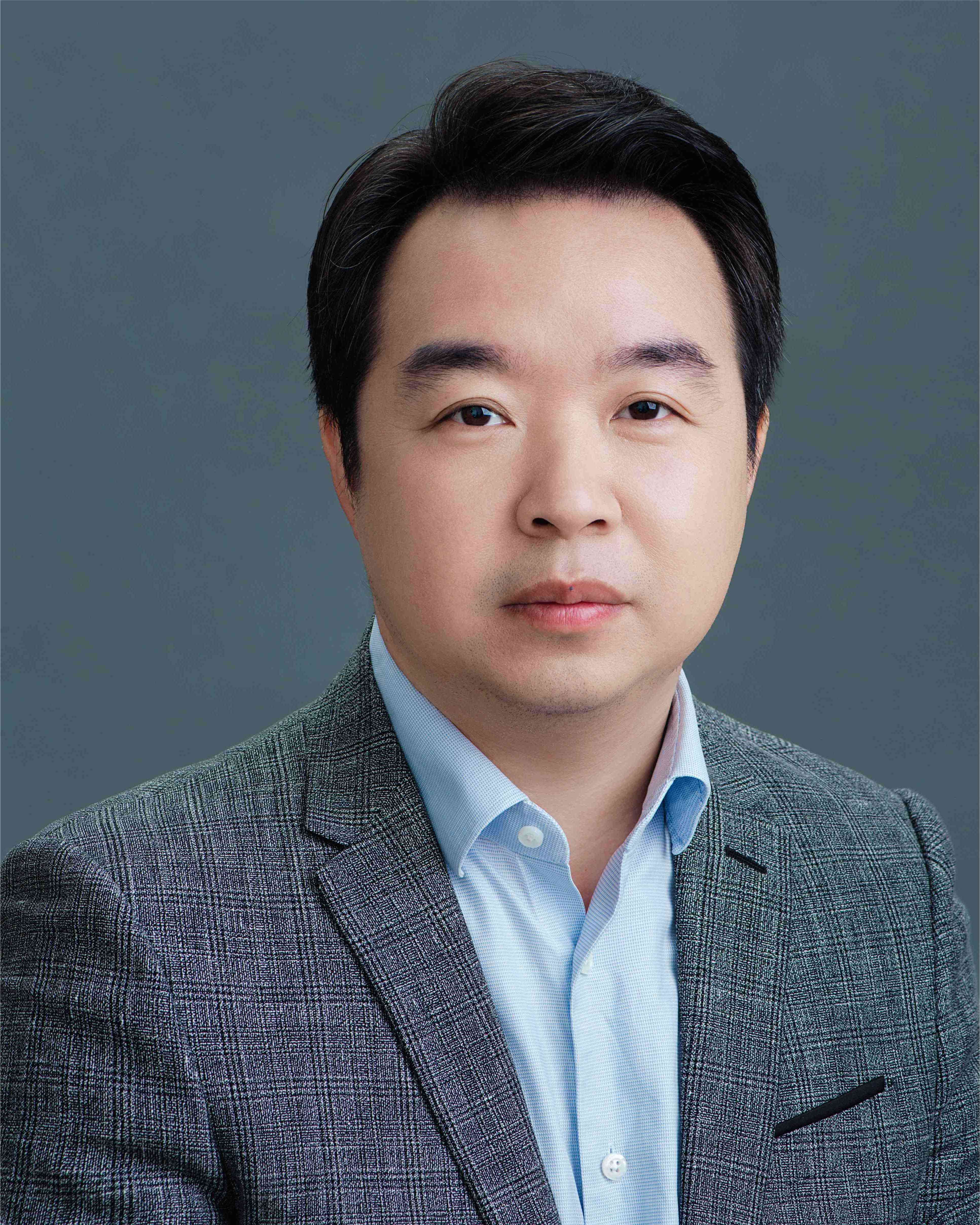}}]{Qinghua Hu} received the B.S., M.S., and Ph.D. degrees from the Harbin Institute of Technology, Harbin, China, in 1999, 2002, and 2008, respectively. After that he joined Department of Computing, The Hong Kong Polytechnical University as a postdoctoral fellow. He became a full professor with Tianjin University in 2012, and now is a Chair Professor and Deputy Dean at College of Intelligence and Computing. His research interest is focused on uncertainty modeling, multi-modality learning, incremental learning and continual learning these years, funded by National Natural Science Foundation of China and The National Key Research and Development Program of China. He has published more than 300 peer-reviewed papers in IEEE TKDE, IEEE TPAMI, IEEE TNNLS, etc. He was a recipient of the best paper award of ICMLC 2015 and ICME 2021. He is an Associate Editor of the IEEE Transactions on Fuzzy Systems, ACTA AUTOMATICA SINICA, and ACTA ELECTRONICA SINICA.
\end{IEEEbiography}
\vspace{-5mm}
\end{document}